\providecommand\FormatOverride{neurips}           %
\providecommand\LogosOverride{ikerlan-cambridge-camlsys-sofair-hitz-flower}  %
\providecommand\TitleBgOverride{camlsys}          %
\PassOptionsToPackage{numbers,square,sort&compress}{natbib}
\documentclass{paper}
\usepackage{colortbl}
\usepackage{pifont}
\usepackage{dsfont}
\usepackage{multirow}
\usepackage{booktabs}
\usepackage{makecell}
\usepackage{wrapfig}
\usepackage{tcolorbox}
\usepackage{enumitem}
\setlist[itemize,enumerate]{itemsep=4pt,parsep=0pt}
\usepackage{ragged2e}
\tcbuselibrary{breakable}
\newcommand{\pmstd}[1]{\,{\tiny$\pm$#1}}
\newcommand{\nostd}{\phantom{\pmstd{0.000}}}     %
\input{preamble_packages}   %
\input{preamble_macros}     %
\input{preamble_adapter}    %
\input{preamble_logos}      %

\newtcolorbox{contributions}{
    breakable,
    colback=blue!4!white,
    colframe=blue!25!white,
    boxrule=0.5pt,
    arc=1mm,
    left=6pt, right=6pt,
    top=6pt, bottom=4pt,
    before skip=8pt,
    after skip=8pt,
    before upper={\textbf{Contributions}\par\smallskip}
}

\newtcolorbox{takeaway}[1][]{
    breakable,
    colback=orange!5!white,
    colframe=orange!50!black,
    boxrule=0.5pt,
    arc=1mm,
    left=6pt, right=6pt,
    top=6pt, bottom=4pt,
    before skip=8pt,
    after skip=8pt,
    #1
}

\paperTitle{DPMA: Learning to Compress Mixture-of-Experts Models}
\paperTitleRunning{DPMA}
\paperKeywords{mixture of experts}
\paperTitleLogo{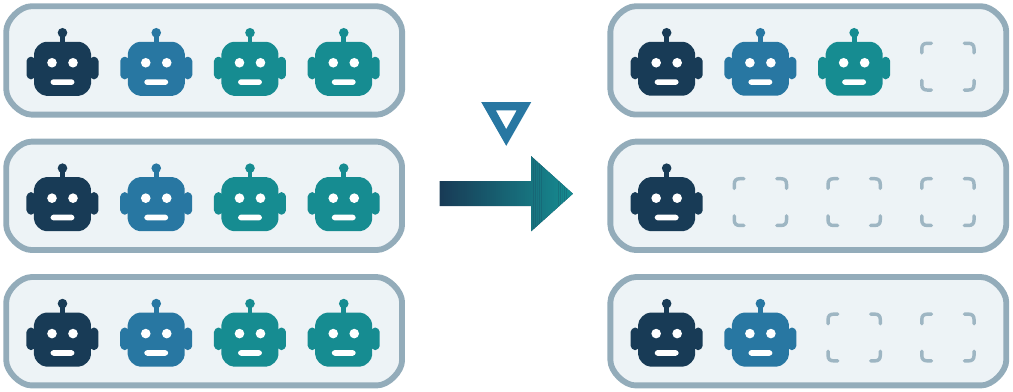}
\paperTitleLogoHeight{4.28em}
\paperTitleLogoWidth{0.20\linewidth}
\paperTitleLogoGap{0.02\linewidth}

\addAffiliation{ikerlan}{Ikerlan Technology Research Center}
\addAffiliation{ehu}{University of the Basque Country UPV/EHU}
\addAffiliation{cam}{University of Cambridge}
\addAffiliation{sofair}{Science of Fundamental AI Research Lab (SOFAIR)}
\addAffiliation{hitz}{HiTZ Basque Center for Language Technology - Ixa, University of the Basque Country UPV/EHU}
\addAffiliation{flower}{Flower Labs}

\addAuthor{Jon Irureta}{ikerlan, ehu, cam}
\addAuthor{Jon Imaz}{ikerlan}
\addAuthor{Philipp Wiesner}{cam, sofair}
\addAuthor{Gorka Azkune}{hitz}
\addAuthor{Javier Fernandez-Marques}{cam, flower}
\addAuthor{Alex Iacob}{cam, sofair}
\addAuthor{Andrej Jovanovi\'c}{cam, sofair}
\addAuthor{Aizea Lojo}{ikerlan}
\addAuthor{Nicholas D. Lane}{cam, sofair, flower}

\correspondingAuthor{Jon Irureta}{jirureta@ikerlan.es}

\begin{document}

\makepapertitle

\section{Introduction}
Many of today's strongest open language models rely on Mixture-of-Experts (MoE) architectures~\citep{jiang2024mixtral, dai2024deepseekmoe, yang2025qwen3}, which allow model capacity to grow keeping computation per token constant.
While dense architectures use the full model for every token, MoE layers route each token to only $k \ll E$ of their $E$ experts~\citep{jacobs1991adaptive, shazeer2017outrageously, fedus2022switch}.

However, this computational advantage does not extend to memory.
All $E$ experts must still remain resident in GPU memory because any expert may be selected for a token~\citep{rajbhandari2022deepspeed}.
MoE compression aims to reduce the memory footprint while preserving model performance.
Many methods do so in a \textit{retraining-free} manner, using a small calibration set to determine which experts to retain or combine without fine-tuning the original model weights~\citep{mc_smoe, diep, hc_moe, reap}.

\begin{figure}[!t]
    \centering
    \includegraphics[width=\linewidth]{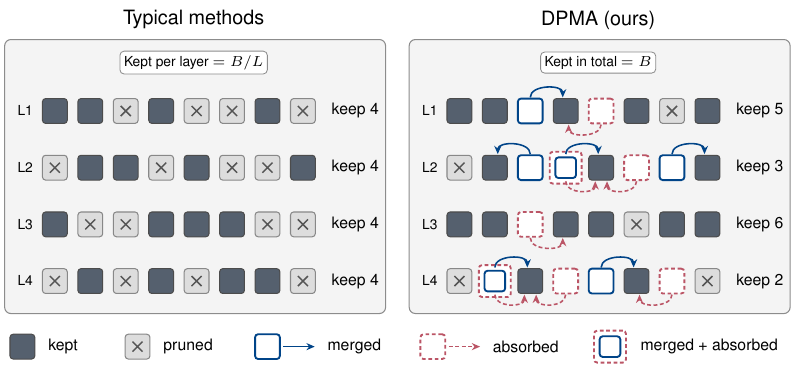}
    \caption{\textbf{\dpma jointly learns compression operations and the allocation of a fixed expert budget.}
    Uniform pruning (left) retains four experts per layer.
    \dpma (right) combines pruning, merging, and absorption while allocating the same total budget of $B=16$ experts across layers.}
    \label{fig:dpma}
\end{figure}

Existing retraining-free methods make multiple restrictive choices in advance: how expert parameters are removed or combined, how routing is adapted when experts are removed, and how the compression budget is distributed across layers.
Most approaches commit upfront to either \emph{pruning} an expert~\citep{reap, diep} or \emph{merging} its parameters into a retained one~\citep{mc_smoe, hc_moe}.
Neither strategy is universally preferable, as the induced errors vary with the setting and cannot be known \textit{a priori}~\citep{reap}.

Furthermore, most methods~\citep{NAEE,EEP,reap, hc_moe, jaiswal2025finding} apply a uniform compression ratio to every layer, regardless of differences in their sensitivity to compression~\citep{diep}.
These decisions together determine how closely the compressed model can reproduce the original model's predictions, suggesting that they could be learned jointly on calibration data.
We therefore ask:

\begin{center}
\itshape
Can gradient descent learn both how and where to compress an MoE model?
\end{center}
\nopagebreak[4]

We introduce \textbf{\dpma} (Differentiable Prune Merge Absorb), which formulates retraining-free MoE compression as a differentiable optimization problem under a global expert budget.
We assign each expert learnable coefficients for retention, parameter merging, and routing contributions.
These coefficients determine whether to \emph{prune} an expert, \emph{merge} its parameters into surviving experts, or \emph{absorb} its routing contribution by forwarding its routing weight while discarding its parameters.
Because merging and absorption act on parameters and routing separately, they can also be combined.
Figure~\ref{fig:dpma} illustrates how \dpma combines these operations while distributing a shared expert budget across layers.

\dpma learns the coefficients and assignments to surviving experts jointly by gradient descent, minimizing the discrepancy between the original and compressed models on calibration data. A budget term drives the total number of retained experts towards a given target. Since this budget is shared by the whole model rather than split evenly across layers, layers compete for it, and each layer's share is learned rather than fixed in advance. A final training stage then retains exactly the target number of experts while continuing to refine the compression.

We make the following contributions:
\begin{enumerate}[leftmargin=1.4em,labelsep=0.4em]
    \item \textbf{Joint optimization of operations and allocation.}
    We introduce \dpma, which represents pruning, merging, and absorption through learnable coefficients and jointly optimizes them with expert retention and assignments to surviving experts.
    The same optimization learns the allocation of operations across layers under a global expert budget.
    \item \textbf{Evaluation across models and budgets.}
    We evaluate \dpma on OLMoE-1B-7B, Qwen1.5-MoE-A2.7B, Qwen3-30B-A3B, and DeepSeek-MoE-16B across eight multiple-choice benchmarks and five compression ratios. Our approach outperform every baseline on each model when accuracy is averaged over benchmarks and compression ratios. 
    \item \textbf{Analysis of learned compression strategies.}
    We show that \dpma learns different mixtures of operations across models and non-uniform allocations across layers.
    Ablations show that allowing all three operations gives higher mean accuracy than every tested restriction, for both variants in zero-shot and one-shot evaluation.
\end{enumerate}
\section{Related Work}\label{sec:background}

Most retraining-free MoE compression methods commit \textit{a priori} to either pruning or merging experts, and usually apply a uniform compression ratio across layers.

\vspace{-3mm}
\paragraph{Expert pruning.}
Expert pruning removes entire experts from the model and restricts routing to those retained.
A simple baseline is frequency-based pruning, which removes the experts selected least often on a calibration set~\citep{jaiswal2025finding}.
\naee~\citep{NAEE} instead searches exhaustively over expert subsets of a prescribed size to minimize the discrepancy between the original and pruned layer outputs.
\reap~\citep{reap} avoids this combinatorial search by ranking experts using their router-weighted output norms, motivated by an upper bound on reconstruction error.
\diep~\citep{diep} formulates expert pruning as a differentiable optimization problem across the network, allowing the number of retained experts to vary between layers under a global budget.

\vspace{-3mm}
\paragraph{Expert merging.}
Expert merging combines the parameters of several experts into a single expert.
\msmoe~\citep{mc_smoe} selects dominant experts using normalized activation frequencies across layers, allowing the number retained to vary between layers.
It groups the remaining experts with dominant experts in the same layer according to the similarity of their router logits.
\hcsmoe~\citep{hc_moe} instead forms a fixed number of groups per layer through hierarchical clustering of experts' mean outputs on calibration data.
Both methods combine the grouped experts through frequency-weighted parameter averaging.
Further approaches refine expert selection and parameter alignment~\citep{jha2026ream}, or merge experts within a shared low-dimensional subspace~\citep{li2026submoe}.

\vspace{-3mm}
\paragraph{Combining operations and preserving routing.}
Some methods combine pruning and merging within a single compression procedure.
\eep~\citep{EEP} uses two sequential evolutionary searches: the first selects retained experts, and the second optimizes separate expert-merging and router-mapping matrices, with a fixed number of experts per layer.
Other work jointly selects pruning and merging decisions through discrete optimization within each layer~\citep{huang2026prime}.
Expert removal can also preserve the original routing slots: removed experts are replaced by constant output vectors estimated from calibration data~\citep{zhang2026mone}, or their tokens are assigned to retained experts while preserving the original routing weights~\citep{yao2026conmoe}.

Unlike prior work, \dpma jointly learns compression operations and layer-wise expert allocation through gradient descent.
Its formulation separates parameter merging from routing contributions, allowing pruning, merging, and absorption to be optimized together under a global expert budget.
Table~\ref{tab:moe_compression_taxonomy} summarizes the main differences between the methods.

\begin{table*}[t]
  \small
  \centering
  \caption{Comparison of selected MoE compression methods in their main configurations by supported operations, per-layer budget allocation, and search strategy.
  Unlike prior work, \dpma jointly learns compression operations and layer-wise expert allocation through gradient descent.}
  \label{tab:moe_compression_taxonomy}
  \begin{tabular}{lccc}
    \toprule
    \textbf{Method} & \textbf{Operations} & \textbf{Per-Layer Budget} & \textbf{Search Strategy} \\
    \midrule
    \freq~\citep{jaiswal2025finding} & Prune & Uniform & Criterion \\
    \naee~\citep{NAEE} & Prune & Uniform & Exhaustive \\
    \reap~\citep{reap} & Prune & Uniform & Criterion \\
    \eep~\citep{EEP} & Prune, then Merge & Uniform & Evolutionary \\
    \diep~\citep{diep} & Prune & Flexible & Gradient Descent \\
    \msmoe~\citep{mc_smoe} & Merge & Flexible & Criterion \\
    \hcsmoe~\citep{hc_moe} & Merge & Uniform & Clustering \\
    \midrule
    \rowcolor{gray!15}
    \dpma (ours) & Prune + Merge + Absorb & Flexible & Gradient Descent \\
    \bottomrule
  \end{tabular}
\end{table*}

\section{Problem Formulation}\label{sec:definition}

\paragraph{MoE layers.}\label{sec:moe}
An MoE layer replaces a transformer's dense feed-forward network with a set of $E$ expert networks, $\mathcal{E}=\{E_1,\ldots,E_E\}$, and a router~\citep{shazeer2017outrageously}.
For an input token representation $\mathbf{x}\in\mathbb{R}^d$, the router produces scores $\mathbf{z}(\mathbf{x})\in\mathbb{R}^E$ and selects the indices of the $k$ highest-scoring experts:
\begin{equation}
    \mathcal{S}(\mathbf{x})=\operatorname{TopK}(\mathbf{z}(\mathbf{x}),k).
\end{equation}
Let $G_i(\mathbf{x})$ denote the weight assigned to expert $i$ by the model's routing rule, including its normalization, with $G_i(\mathbf{x})=0$ for $i\notin\mathcal{S}(\mathbf{x})$.
The routed output is
\begin{equation}
    \operatorname{MoE}(\mathbf{x},\mathcal{E},G)=\sum_{i\in\mathcal{S}(\mathbf{x})}G_i(\mathbf{x})E_i(\mathbf{x}).
\end{equation}
Only the selected experts are evaluated, with $k$ fixed by the model architecture~\citep{lepikhin2021gshard, fedus2022switch}.
Where present, always-active shared experts contribute separately and are excluded from $\mathcal{E}$.

\paragraph{Compression objective.}
Consider a pre-trained model with $L$ MoE layers, each containing $E$ routed experts.
Given a target compression ratio $c\in(0,1)$, we seek a compressed model retaining a total of $B=\lfloor(1-c)LE\rfloor$ expert networks across its MoE layers.
Shared experts and non-expert parameters are excluded from the budget.
Let $n_\ell$ denote the number of retained expert networks in layer $\ell$, and $m_\ell$ its prescribed minimum.
The allocation must satisfy
\begin{equation}
    \sum_{\ell=1}^{L}n_\ell=B,
    \qquad m_\ell\leq n_\ell\leq E.
\end{equation}
Subject to these constraints, we seek a compression recipe that minimizes the discrepancy between the original and compressed models on a calibration dataset $\mathcal{X}$, while keeping the original model parameters frozen.

\section{Method}\label{sec:methods}
\dpma represents expert retention, parameter merging, and routing contributions through learnable coefficients, and jointly optimizes these coefficients and assignments to surviving experts by gradient descent under the budget constraints of Section~\ref{sec:definition}.
Figure~\ref{fig:dpma} illustrates how the resulting compression recipe combines operations and allocates the expert budget across layers.

\subsection{Pruning, merging, and absorption}\label{subsec:operations}
Removing an expert from a layer raises two separate questions: what becomes of its \textit{parameters}, and what becomes of its \textit{routing slot}, i.e., the entry through which the router scores and selects it.
\dpma considers three operations:
\begin{itemize}[leftmargin=*,labelsep=0.5em]
    \item \textbf{Prune.}
    The expert's parameters are discarded and its routing slot is masked to prevent selection.
    \item \textbf{Merge.}
    The expert's parameters are combined with those of one or more retained experts, after which its own parameters are discarded.
    Its routing slot is masked, as in pruning.
    \item \textbf{Absorb.}
    The expert's parameters are discarded, but its routing slot remains selectable.
    When the slot is selected, a learned fraction of its routing weight is forwarded to one or more retained experts, which process the token in its place.
\end{itemize}
Merging and absorption act on parameters and routing separately, so they can be combined for the same removed expert, as illustrated in Figure~\ref{fig:dpma}.

\subsection{Parameterizing the compressed model}
\begin{wraptable}{r}{0.3\textwidth}
\centering
\vspace{-4mm}
\caption{Operations based on $\alpha$ and $\beta$ values.}
\small
\begin{tabular}{@{}ccl@{}}
\toprule
$\alpha_e$ & $\beta_e$ & Operation\\
\midrule
$0$ & $0$ & Prune\\
$0$ & $\neq 0$ & Merge\\
$>0$ & $0$ & Absorb\\
$>0$ & $\neq 0$ & Merge + absorb\\
\bottomrule
\end{tabular}
\label{tab:operations}
\vspace{1.5em}
\end{wraptable}
\label{sec:learning}
We describe a single MoE layer, omitting the layer index, and write $\theta_i$ for the parameters of expert~$E_i$.
We assign each expert a presence coefficient $\rho_i\in[0,1]$, an absorption coefficient $\alpha_i\in[0,1]$, and a merging coefficient $\beta_i\in\mathbb{R}$.

\paragraph{Encoding the operations.}
We express the compression operations through two independently weighted contributions: one to a retained expert's parameters and one to its routing weight.
Consider a removed expert $E_e$, with $\rho_e=0$, assigned to a retained expert $E_r$, with $\rho_r=1$.
For given routing weights $G_e(\mathbf{x})$ and $G_r(\mathbf{x})$, the two experts contribute
\begin{equation}
    G_r(\mathbf{x})E_r(\mathbf{x};\theta_r)+G_e(\mathbf{x})E_e(\mathbf{x};\theta_e).
\end{equation}
Merging and absorption replace these contributions with $\tilde G_r(\mathbf{x})E_r(\mathbf{x};\tilde\theta_r)$, where
\begin{equation}
    \tilde G_r(\mathbf{x})=G_r(\mathbf{x})+\alpha_e\,G_e(\mathbf{x}),
    \qquad
    \tilde\theta_r=\theta_r+\beta_e\,(\theta_e-\theta_r).
    \label{equ:modified_router_weights}
\end{equation}

The \emph{absorption weight} $\alpha_e\in[0,1]$ controls the fraction of the removed expert’s routing weight forwarded to $E_r$.
The \emph{merging weight} $\beta_e\in\mathbb{R}$ moves $E_r$ a fraction $\beta_e$ of the way toward $E_e$ in parameter space, with negative values moving it away.
Table~\ref{tab:operations} summarizes the operations obtained by enabling or disabling these contributions.

\paragraph{Routing mask.}
The routing mask controls which expert slots remain selectable after compression.
We adjust the router logits before selection to suppress slots of pruned experts and experts merged without absorption, while preserving slots that forward routing weight.
For each slot $i$, we define
\begin{equation}
    m_i=\max(\rho_i,\ \alpha_i,\ \epsilon),
    \qquad
    \tilde z_i(\mathbf{x})=z_i(\mathbf{x})+\log m_i
    \label{eq:mask}
\end{equation}
where $\epsilon\in(0,1)$ is a small constant that keeps the logarithm defined.
A retained expert has $\rho_i=1$, so its logit is unchanged.
For a removed expert with $\rho_i=\alpha_i=0$, the logit is shifted by $\log\epsilon$, suppressing selection as $\epsilon\to0$.
If $\rho_i=0$ and $\alpha_i>0$, the slot remains selectable with a logit adjustment of $\log\max(\alpha_i,\epsilon)$.
Applying the model's own selection and normalization rule to the modified logits yields the selected index set $\mathcal{S}(\mathbf{x})$ and routing weights $G(\mathbf{x})$.

\paragraph{Assignment matrix.}
The assignment matrix determines which retained experts receive each removed expert's parameter and routing contributions.
We learn these assignments through continuous weights.
We define $A\in[0,1]^{E\times E}$, where $A_{er}$ specifies the share of expert $E_e$'s contributions assigned to expert $E_r$.
Each row sums to one and $A_{ee}=0$ excludes self-assignment.
Before optimization, we compute a fixed \emph{signature} $u_i\in\mathbb{R}^{d}$ for each expert using the uncompressed model.
The signature is the mean of that expert's output vectors over the calibration tokens routed to it, normalized to unit $\ell_2$ norm.
We compare these signatures through a learned metric $M=I+VV^{\top}$, where $V\in\mathbb{R}^{d\times8}$ is shared across all layers.
With a temperature $\tau>0$, the assignment weights are
\begin{equation}
    A_{er}=\frac{\rho_r\exp\!\bigl(u_e^{\top}M\,u_r/\tau\bigr)}
    {\sum_{r'\neq e}\rho_{r'}\exp\!\bigl(u_e^{\top}M\,u_{r'}/\tau\bigr)},
    \qquad A_{ee}=0.
    \label{eq:assign}
\end{equation}

The temperature $\tau$ controls how concentrated each row is, with lower values favouring the highest-scoring recipients.
The factor $\rho_r$ down-weights recipients with intermediate presence and excludes those with $\rho_r=0$.
We recompute $A$ at every forward pass using the current metric and presence coefficients.
Its entries are not independently trained parameters.

\paragraph{Constructing the compressed layer.}
We now combine the routing mask, assignment weights, and operation coefficients to compute the compressed layer's output.
For recipient $r$, the two-expert construction in Equation~\eqref{equ:modified_router_weights} extends to
\begin{align}
    \tilde G_r(\mathbf{x})&=\rho_r\,G_r(\mathbf{x})
    +\sum_{e\neq r}(1-\rho_e)\,\alpha_e\,A_{er}\,G_e(\mathbf{x}),
    \label{eq:absorb}\\
    \tilde\theta_r&=\theta_r+\sum_{e}(1-\rho_e)\,\beta_e\,A_{er}\,(\theta_e-\theta_r).
    \label{eq:merge}
\end{align}
The factor $1-\rho_e$ enables parameter merging and routing-weight forwarding from a removed expert ($\rho_e=0$).
It is zero for a retained expert ($\rho_e=1$), whose parameters and routing weight are not redistributed to other experts.
If the incoming merging coefficients $(1-\rho_e)\beta_e A_{er}$ in Equation~\eqref{eq:merge} sum to more than one, we renormalize them to prevent the coefficient on the retained expert's original parameters $\theta_r$ from becoming negative.
The combined routing weights $\tilde G(\mathbf{x})$ depend on the token, while the merged parameters $\tilde\theta$ do not.

After applying Equtions~\eqref{eq:absorb} and \eqref{eq:merge}, the compressed layer output is
\begin{equation}
    \sum_{r=1}^{E}\tilde G_r(\mathbf{x})\,E_r(\mathbf{x};\tilde\theta_r).
    \label{eq:output}
\end{equation}
Note that absorption can activate retained experts outside $\mathcal{S}(\mathbf{x})$ when they receive routing weight from a selected slot belonging to a removed expert.

\subsection{Gradient-based optimization}\label{sec:optimisation}
We jointly optimize the operation coefficients and expert assignments by gradient descent while keeping the original model parameters frozen.
Optimization proceeds in two stages: 
a soft stage with continuous presence coefficients followed by a hard stage with binary selection of exactly B experts.

\paragraph{Calibration loss.}
The calibration loss encourages the compressed model to match the original model's predictions and MoE-layer outputs.
During the initial soft training stage, we combine this loss with the budget term $\mathcal{L}_{\mathrm{budget}}$, defined below:

\begin{equation}
    \mathcal{L}=
    \mathrm{KL}\bigl(p_{\mathrm{orig}}\Vert p_{\mathrm{comp}}\bigr)
    +\eta\sum_{l=1}^{L}
    \frac{\sum_{\mathbf{x}\in\mathcal{X}_l}
    \bigl\lVert\widetilde{\mathrm{MoE}}_l(\mathbf{x})-\mathrm{MoE}_l(\mathbf{x})\bigr\rVert_2^2}
    {\sum_{\mathbf{x}\in\mathcal{X}_l}\bigl\lVert\mathrm{MoE}_l(\mathbf{x})\bigr\rVert_2^2}
    +\mathcal{L}_{\mathrm{budget}}.
    \label{eq:obj}
\end{equation}

Here, $p_{\mathrm{orig}}$ and $p_{\mathrm{comp}}$ are the original and compressed models' next-token distributions on the calibration data.
The KL term encourages the compressed model to preserve the original model's predictions.
The reconstruction term additionally encourages agreement between their MoE-layer outputs on inputs $\mathcal{X}_l$, with normalization preventing layers with larger outputs from dominating. Refer to Appendix~\ref{app:ablations-sensitivity} for further details about the impact of the terms.

\paragraph{Hard-concrete gates.}
We use hard-concrete gates~\citep{louizos2018learning} to optimize continuous coefficients while allowing exact zeros that disable expert presence, absorption, or merging.
Each gate takes values in $[0,1]$, while a learned scale $b\in\mathbb{R}$ allows signed, unbounded merging coefficients.
We jointly optimize the gate logits $\phi^{\rho}$, $\phi^{\alpha}$, and $\phi^{\beta}$, merging scales $b$, and shared metric factor $V$, with their mappings given in Table~\ref{tab:params}.
Appendix~\ref{app:pnz} specifies the gate construction and deterministic evaluation.

\begin{table}[h]
\centering
\caption{Learned parameters and derived quantities of \dpma.
The $4LE+8d$ learned parameters comprise four scalars per expert and a shared metric factor $V\in\mathbb{R}^{d\times8}$.}
\small
\renewcommand{\arraystretch}{1.2}
\begin{tabular}{@{}p{1.5cm}p{4.8cm}p{1.8cm}p{3.3cm}@{}}
\toprule
\textbf{Learned} & \textbf{Derived at each forward pass} & \textbf{Range} & \textbf{Role}\\
\midrule
$\phi^{\rho}$ & $\rho=\mathrm{gate}(\phi^{\rho})$ & $[0,1]$ & Presence\\
$\phi^{\alpha}$ & $\alpha=\mathrm{gate}(\phi^{\alpha})$ & $[0,1]$ & Absorption weight\\
$\phi^{\beta},\,b$ & $\beta=b\cdot\mathrm{gate}(\phi^{\beta})$ & $\mathbb{R}$ & Merging weight\\
$V$ & $A_{er}\propto\rho_r\,e^{\,u_e^{\top}(I+VV^{\top})u_r/\tau}$ & $\sum_r A_{er}{=}1$ & Assignment\\
\bottomrule
\end{tabular}
\label{tab:params}
\end{table}

\paragraph{Budget constraint.}
During the soft stage, we constrain the expected number of experts with non-zero presence to the global budget $B$.
For $P_{l,i}=\Pr(\rho_{l,i}\neq0)$, the expected count $\sum_{l,i}P_{l,i}$ is differentiable with respect to the presence logits.
We penalize its deviation from $B$ using an augmented Lagrangian:

\begin{equation}
    \bar h=\frac{1}{LE}\Bigl(\sum_{l,i}P_{l,i}-B\Bigr),
    \qquad
    \mathcal{L}_{\mathrm{budget}}=\mu\,\bar h+\tfrac{\kappa}{2}\bar h^{2},
    \qquad
    \mu\leftarrow\mu+\eta_\mu\,\bar h.
    \label{eq:lagrangian}
\end{equation}
Here, $\bar h$ is the normalized budget deviation, $\kappa$ controls the quadratic penalty, and $\mu$ is a dual variable with ascent step size $\eta_\mu$.
We hold $\mu$ fixed during each gradient-descent update, then recompute $\bar h$ and update $\mu$.
The global constraint allows the calibration loss to guide how the budget is allocated across layers.
Appendix~\ref{app:pnz} gives the closed-form gate probabilities and their derivation.

\paragraph{Training stages.}
We train in two stages.
In the soft stage, we optimize the full objective with presence coefficients in $[0,1]$.
However, constraining the expected number of retained experts does not guarantee a binary selection of exactly $B$ experts.
In the hard stage, we set the presence coefficients to binary values, selecting the $B$ experts with the highest presence probabilities $P$, subject to retaining at least $k+2$ experts per layer.
The presence logits continue to learn through a straight-through estimator~\citep{bengio2013estimating}:
\begin{equation}
    \rho=\rho_{\mathrm{hard}}+\bigl(\rho_{\mathrm{soft}}-\mathrm{sg}[\rho_{\mathrm{soft}}]\bigr),
    \label{eq:st}
\end{equation}
where $\mathrm{sg}$ stops the gradient through its argument.
The forward pass uses the binary selection $\rho_{\mathrm{hard}}$, while the backward pass uses gradients through the relaxed coefficients $\rho_{\mathrm{soft}}$.
The selected experts can change during training, while their total count remains $B$.
Operation coefficients and assignments continue to be optimized.
We set $\eta=0$ and optimize only the prediction KL loss.
The budget penalty is no longer needed because the binary selection enforces the constraint.
Stage durations and optimization schedules are specified in Appendix~\ref{app:hyperparameters}.

\subsection{Hard and soft assignments}
To offer different trade-offs between accuracy and the number of active experts, we introduce two variants of \dpma.
Both use the same $B$ retained experts from a single optimization run and differ in how they assign removed experts' parameter and routing contributions.
We compute the merged parameters once and discard the removed experts' parameter blocks, while preserving routing slots used for absorption.

\dpmahard assigns each removed expert to its highest-weight recipient by setting the largest entry in its row of $A$ to one and all others to zero.
Each selected routing slot then contributes to at most one retained expert, so a token activates at most $k$ routed experts per layer.

\dpmasoft keeps the three largest entries in each removed expert's row of $A$, sets the others to zero, and renormalizes the retained weights to sum to one.
A selected slot can therefore distribute its routing weight across several recipients, potentially activating more than $k$ routed experts per token.

\section{Evaluation}\label{sec:experimental_Setup}

We evaluate \dpma across four MoE models, eight multiple-choice benchmarks, and five compression ratios.
Our main findings are:
\begin{takeaway}[colback=yellow!8!white,colframe=yellow!40!orange]
\begin{itemize}[leftmargin=*,labelsep=0.5em,topsep=0pt]
    \item \textbf{\dpma reaches higher mean accuracy across all four models.}
    Both \dpma variants achieve higher accuracy than every baseline, averaged over benchmarks and compression ratios.
    At $50\%$ expert removal, \dpmasoft outperforms every baseline on all four models.
    At $75\%$ removal, both variants do so, with gains of up to $6.7$ percentage points (Section~\ref{sec:results}).
     \item \textbf{\dpma learns model-specific compression recipes by combining operations.}
    \dpma tailors layer-wise budgets and operation mixes to each model (Section~\ref{sec:compression-recipes}), and combining pruning, merging, and absorption outperforms any tested restriction (Section~\ref{sec:evaluation-ablations}).
    \item \textbf{\dpma's accuracy advantage persists at matched expert usage.}
    Both variants retain their lead when every method evaluates the same number of experts per token per layer (Section~\ref{sec:active-expert-counts}).
\end{itemize}
\end{takeaway}

\subsection{Experimental setup}\label{sec:evaluation-setup}

\paragraph{Models and budgets.}
We evaluate OLMoE-1B-7B, Qwen1.5-MoE-A2.7B, Qwen3-30B-A3B, and DeepSeek-MoE-16B (Table~\ref{tab:moe-architectures}) at compression ratios $c\in\{0.25,0.40,0.50,0.60,0.75\}$.
The ratio specifies the fraction of routed experts removed, excluding shared experts and non-expert parameters from the budget.
\dpma retains at least $k+2$ experts in each layer, where $k$ is its original number of active routed experts.

\begin{table}[b]
  \centering
  \small
  \setlength{\tabcolsep}{5pt}
  \caption{The four MoE models.
  Experts are per MoE layer; active experts are the routed experts each token uses (top-$k$).
  Active parameters per token count every non-expert parameter (attention, embeddings, output layer, routers), the shared expert(s), and $k$ routed experts per layer.}
  \label{tab:moe-architectures}
  \begin{tabular}{lrrrlrr}
    \toprule
    Model & Layers & \makecell[r]{Experts\\per layer} & \makecell[r]{Active\\experts} &
      \makecell[l]{Shared\\experts} & \makecell[r]{Total\\parameters} & \makecell[r]{Active\\parameters} \\
    \midrule
    OLMoE-1B-7B       & 16 & 64 & 8 & none               & 7B  & 1B \\
    Qwen1.5-MoE-A2.7B & 24 & 60 & 4 & 1 per layer & 14.3B & 2.7B \\
    Qwen3-30B-A3B     & 48 & 128 & 8 & none              & 30.5B & 3B \\
    DeepSeek-MoE-16B & 28 & 64  & 6 & 2 per layer & 16.4B & 2.8B \\
    \bottomrule
  \end{tabular}
\end{table}

\vspace{-2mm}
\paragraph{Baselines.}
We compare \dpmahard and \dpmasoft against random pruning, frequency-based pruning~\citep{jaiswal2025finding}, \reap~\citep{reap}, \diep~\citep{diep}, \msmoe~\citep{mc_smoe}, and \hcsmoe~\citep{hc_moe}, described in Section~\ref{sec:background}.
In our evaluated configurations, \diep and \msmoe allocate experts non-uniformly across layers, while the remaining baselines use uniform compression.

\vspace{-2mm}
\paragraph{Calibration and optimization.}
We use the same C4~\citep{c4} calibration set for all methods: 32 sequences of 1{,}024 tokens, or 128 sequences for Qwen3.
We run each \dpma configuration with three seeds.
We select the checkpoint with the lowest KL divergence from the original model on a separate, held-out C4 set.
Appendix~\ref{app:hyperparameters} specifies the optimization settings.

\vspace{-2mm}
\paragraph{Benchmarks and metrics.}
We evaluate the original and compressed models on ARC-Challenge and ARC-Easy~\citep{arc}, BoolQ~\citep{boolq}, HellaSwag~\citep{hellaswag}, MMLU~\citep{mmlu}, OpenBookQA~\citep{bookqa}, PIQA~\citep{piqa}, and WinoGrande~\citep{winogrande}.
The main comparison reports mean zero-shot accuracy across these eight benchmarks, with an additional mean over the five compression ratios.
Where reported, standard deviations describe variation across optimization seeds or random pruning draws.

\begin{widetable}[t]
\centering
\fontsize{8}{9}\selectfont
\renewcommand{\pmstd}[1]{\,{\scriptsize$\pm$#1}}
\renewcommand{\nostd}{\phantom{\pmstd{0.0}}}
\setlength{\tabcolsep}{5pt}
\caption{\textbf{Both \dpma variants outperform every baseline in mean accuracy on all four models.}
Scores are mean zero-shot accuracy (\%) over eight multiple-choice benchmarks (random guessing: $34.38\%$).
Compression ratios specify the fraction of routed experts removed.
The Mean column averages over the five ratios.
$\pm$ is the standard deviation across seeds or random draws.
Bold and underlining mark the best and second-best compressed results per model and column.}
\label{tab:main-results-0shot}
\begin{tabular}{llcccccc}
\toprule
 & & \multicolumn{5}{c}{\textbf{Compression ratio}} & \\
\cmidrule(lr){3-7}
\textbf{Model} & \textbf{Method} & \textbf{25\%} & \textbf{40\%} & \textbf{50\%} & \textbf{60\%} & \textbf{75\%} & \textbf{Mean} \\
\midrule
\multirow{8}{*}{\makecell[l]{\textbf{OLMoE-1B-7B}\\Uncompressed: 65.3\%}}
 & \rand    & 55.2\pmstd{1.0} & 47.5\pmstd{1.7} & 42.1\pmstd{1.1} & 37.9\pmstd{1.0} & 34.0\pmstd{0.7} & 43.3 \\
 & \freq & 54.5\nostd & 45.5\nostd & 41.3\nostd & 36.5\nostd & 33.9\nostd & 42.3 \\
 & \reap      & \textbf{58.8}\nostd & 49.7\nostd & 44.7\nostd & 39.1\nostd & 33.7\nostd & 45.2 \\
 & \diep      & 54.4\pmstd{4.5} & 45.5\pmstd{2.5} & 41.4\pmstd{0.7} & 37.6\pmstd{1.2} & 34.8\pmstd{0.7} & 42.8 \\
 & \msmoe    & 49.2\nostd & 41.0\nostd & 36.4\nostd & 34.5\nostd & 34.2\nostd & 39.1\\
 & \hcsmoe   & \underline{58.0}\nostd & 50.2\nostd & 45.1\nostd & 41.1\nostd & 36.4\nostd & 46.2 \\
\cmidrule(lr){2-8}
 & \dpmahard & 57.1\pmstd{0.7} & \underline{51.5}\pmstd{1.2} & \underline{46.8}\pmstd{0.4} & \underline{43.0}\pmstd{2.2} & \underline{38.9}\pmstd{0.1} & \underline{47.4} \\
 & \dpmasoft & 57.9\pmstd{0.5} & \textbf{52.1}\pmstd{1.3} & \textbf{46.9}\pmstd{1.0} & \textbf{43.5}\pmstd{1.8} & \textbf{39.5}\pmstd{0.8} & \textbf{48.0} \\
\midrule
\multirow{8}{*}{\makecell[l]{\textbf{Qwen1.5-MoE-A2.7B}\\Uncompressed: 65.8\%}}
 & \rand    & 55.1\pmstd{8.7} & 50.4\pmstd{7.5} & 48.5\pmstd{6.1} & 43.4\pmstd{4.1} & 38.6\pmstd{2.6} & 47.2 \\
 & \freq & 56.5\nostd & 54.7\nostd & 51.3\nostd & 47.3\nostd & 37.5\nostd & 49.5 \\
 & \reap      & 61.6\nostd & 58.4\nostd & 54.4\nostd & 49.8\nostd & 41.8\nostd & 53.2 \\
 & \diep      & 60.8\pmstd{0.7} & 55.7\pmstd{0.9} & 52.5\pmstd{0.5} & 46.0\pmstd{2.9} & 37.7\pmstd{0.9} & 50.6 \\
 & \msmoe    & 51.4\nostd & 42.8\nostd & 40.5\nostd & 38.2\nostd & 36.8\nostd & 41.9 \\
 & \hcsmoe   & 61.9\nostd & 58.3\nostd & 55.1\nostd & 52.3\nostd & 46.8\nostd & 54.9 \\
\cmidrule(lr){2-8}
 & \dpmahard & \underline{62.8}\pmstd{0.2} & \underline{58.8}\pmstd{0.5} & \underline{55.5}\pmstd{1.1} & \underline{52.5}\pmstd{1.0} & \underline{47.5}\pmstd{0.3} & \underline{55.4}\\
 & \dpmasoft & \textbf{63.3}\pmstd{0.2} & \textbf{59.2}\pmstd{0.7} & \textbf{56.4}\pmstd{1.0} & \textbf{53.4}\pmstd{1.0} & \textbf{49.1}\pmstd{0.7} & \textbf{56.3} \\
\midrule
\multirow{8}{*}{\makecell[l]{\textbf{Qwen3-30B-A3B}\\Uncompressed: 72.0\%}}
 & \rand    & 49.7\pmstd{11.2} & 40.7\pmstd{3.9} & 37.1\pmstd{1.2} & 36.7\pmstd{0.8} & 34.6\pmstd{0.5} & 39.7 \\
 & \freq & 71.0\nostd & 67.4\nostd & 56.5\nostd & 47.9\nostd & 44.8\nostd & 57.5 \\
 & \reap      & 71.0\nostd & 66.6\nostd & 61.4\nostd & 55.8\nostd & 42.8\nostd & 59.5 \\
 & \diep      & 69.1\pmstd{0.4} & 61.2\pmstd{2.6} & 50.2\pmstd{11.1} & 40.2\pmstd{6.8} & 35.7\pmstd{0.8} & 51.3 \\
 & \msmoe    & 70.1\nostd & 64.7\nostd & 41.8\nostd & 38.9\nostd & 35.9\nostd & 50.3 \\
 & \hcsmoe   & 63.8\nostd & 52.3\nostd & 45.7\nostd & 40.3\nostd & 37.2\nostd & 47.9 \\
\cmidrule(lr){2-8}
 & \dpmahard & \underline{71.1}\pmstd{0.1} & \underline{68.3}\pmstd{0.3} & \underline{65.6}\pmstd{0.1} & \underline{60.5}\pmstd{1.4} & \underline{50.0}\pmstd{1.9} & \underline{63.1} \\
 & \dpmasoft & \textbf{71.3}\pmstd{0.1} & \textbf{68.5}\pmstd{0.7} & \textbf{65.8}\pmstd{0.3} & \textbf{60.9}\pmstd{1.2} & \textbf{51.5}\pmstd{0.6} & \textbf{63.6} \\
 \midrule
\multirow{8}{*}{\makecell[l]{\textbf{DeepSeek-MoE-16B}\\Uncompressed: 62.4\%}}
 & \rand    & 58.0\pmstd{0.7} & 46.8\pmstd{10.0} & 48.7\pmstd{0.5} & 38.2\pmstd{5.9} & 35.5\pmstd{3.0} & 45.4 \\
 & \freq & 59.8\nostd & 57.4\nostd & \underline{55.0}\nostd & 49.5\nostd & 40.5\nostd & 52.4 \\
 & \reap      & \textbf{61.4}\nostd & 57.4\nostd & 53.5\nostd & 48.0\nostd & 41.1\nostd & 52.3 \\
 & \diep      & 54.2\pmstd{1.9} & 49.4\pmstd{2.1} & 45.7\pmstd{1.2} & 39.0\pmstd{4.6} & 34.1\pmstd{0.6} & 44.5 \\
 & \msmoe    & 57.9\nostd & 53.5\nostd & 48.5\nostd & 33.4\nostd & 33.9\nostd & 45.4 \\
 & \hcsmoe   & 59.7\nostd & 55.8\nostd & 52.6\nostd & 47.1\nostd & 39.8\nostd & 51.0 \\
\cmidrule(lr){2-8}
 & \dpmahard & 61.0\pmstd{0.3} & \underline{57.5}\pmstd{0.1} & 54.9\pmstd{0.3} & \underline{51.6}\pmstd{0.8} & \underline{44.1}\pmstd{1.4} & \underline{53.8} \\
 & \dpmasoft & \underline{61.1}\pmstd{0.2} & \textbf{57.7}\pmstd{0.1} & \textbf{55.5}\pmstd{0.4} & \textbf{52.3}\pmstd{0.4} & \textbf{44.8}\pmstd{1.1} & \textbf{54.3} \\
\bottomrule
\end{tabular}
\end{widetable}

\subsection{Accuracy across models and compression ratios}\label{sec:results}

\dpma achieves higher accuracy than the evaluated baselines across a broad range of models and compression ratios (Table~\ref{tab:main-results-0shot}).
\dpmasoft outperforms every baseline in $18$ of the $20$ combinations of model and compression ratio.
Both variants outperform every baseline on each model when averaged over the eight benchmarks and five compression ratios.
Compared with the strongest baseline for each model, these averages improve by $0.5$--$3.6$ percentage points for \dpmahard and $1.4$--$4.1$ points for \dpmasoft.

In contrast to \dpma’s consistency, baseline performance is brittle across both architectures and compression regimes. No single baseline transfers reliably across models: \hcsmoe leads on OLMoE and Qwen1.5-MoE, \reap on Qwen3, and \freq on DeepSeek-MoE, yet \hcsmoe ranks behind every non-random baseline on Qwen3. Rankings also invert as compression tightens—for instance, \msmoe performs competitively at low compression on Qwen3 and DeepSeek-MoE before collapsing below random pruning at higher ratios, and \reap’s early advantage on OLMoE and Qwen3 disappears at $75\%$ removal.

\dpma's advantage is more pronounced at higher compression ratios. At $25\%$ expert removal, \reap achieves the highest accuracy on OLMoE and DeepSeek-MoE. At $50\%$, \dpmasoft outperforms every baseline on all four models, by $0.5$--$4.4$ percentage points. \dpmahard does so on three models and falls $0.1$ points below frequency-based pruning on DeepSeek-MoE. At $75\%$ removal, both variants outperform every baseline on all four models. The largest gains occur on Qwen3, where \dpmahard and \dpmasoft achieve $50.0\%$ and $51.5\%$ accuracy, respectively, compared with $44.8\%$ for the strongest baseline, frequency-based pruning.

\subsection{How \dpma allocates experts and selects operations}\label{sec:compression-recipes}
\begin{figure}
    \centering
    \includegraphics[width=\linewidth]{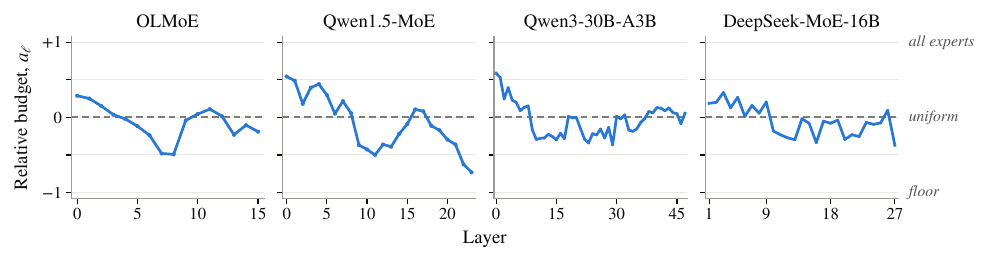}
    \vspace{-5mm}
    \caption{\textbf{\dpma learns where to compress, moving the expert budget away from a uniform split.}
    Relative budget $a_\ell$ of each layer: $0$ is the uniform split, $+1$ keeps all of the layer's experts, and $-1$ cuts it to \dpma's minimum.}
    \label{fig:expert_budget_allocation}
\end{figure}

Unlike alternative approaches, \dpma jointly learns both the compression operations and the per-layer budget allocation, and we attribute its consistency across models and compression regimes to this flexibility. To understand these learned strategies, we analyze how \dpma distributes the expert budget across layers and selects compression operations. Across all models, \dpma generally retains more experts in early layers than a uniform allocation would, while compressing intermediate layers more strongly (Figure~\ref{fig:expert_budget_allocation}).
The final layers are also compressed more strongly on Qwen1.5-MoE and DeepSeek-MoE, particularly on Qwen1.5-MoE. Figure~\ref{fig:budet-split-all-methods} in the appendix compares these allocations with those of \msmoe and \diep at each compression ratio. \msmoe often stays closer to a uniform allocation, whereas \diep shows sharper changes between neighbouring layers and greater variation across seeds.

The mixture of operations differs across models (Table~\ref{tab:operations-chosen-compression} in the appendix). At $50\%$ expert removal, \dpma prunes $85.9\%$ of the removed experts on Qwen3, but only $8.4\%$ on OLMoE, where absorption, alone or together with merging, is used for $84.2\%$ of them; Qwen1.5-MoE and DeepSeek-MoE likewise combine pruning, merging, and absorption. Whereas each baseline is committed to a single operation, \dpma selects the mixture for each model, ranging from nearly pure pruning to predominantly absorption. This mixture is not dictated by which operation works best among the baselines: although a pruning method achieves the highest baseline mean accuracy on DeepSeek-MoE, \dpma relies largely on merging and absorption there and still achieves higher mean accuracy.

\subsection{Accuracy gains matching active experts}\label{sec:active-expert-counts}
\dpmahard achieves higher mean accuracy than every baseline while activating no more experts per token.
\dpmasoft achieves even higher accuracy, but can activate additional experts (Appendix~\ref{app:hard-vs-soft}).
To assess whether its advantage over the baselines depends on these additional experts, we compare all methods with the same number of active experts per token.

\begin{figure}[t]
\centering
\begin{minipage}[t]{0.48\linewidth}
  \centering
  \includegraphics[width=\linewidth]{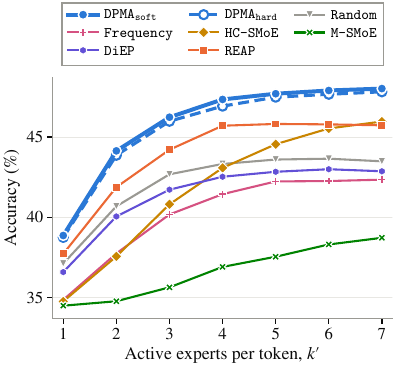}
  \caption{\textbf{Both \dpma variants outperform every baseline when the number of active experts is matched.} Mean zero-shot accuracy over the eight multiple-choice benchmarks on OLMoE, averaged over the five compression ratios, when every method evaluates exactly $k'$ experts per token per layer.}
  \label{fig:fixed-flops}
\end{minipage}\hfill
\begin{minipage}[t]{0.48\linewidth}
  \centering
  \includegraphics[width=\linewidth]{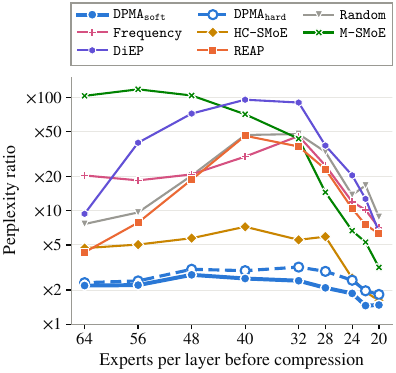}
  \caption{\textbf{\dpma limits the perplexity increase from compression even when fewer experts are available.} We randomly remove experts from OLMoE to obtain models with $20$--$64$ experts per layer, then compress each model by $50\%$. Ratios compare WikiText-2 perplexity after and before compression, using the geometric mean over three nested draws.}
  \label{fig:ppl-ratio-expert-count}
\end{minipage}
\end{figure}

On OLMoE, we require every method to evaluate exactly $k'$ experts per token in each layer, for $k'\in\{1,\ldots,7\}$, instead of the original $k=8$. For the baselines, we set the router's top-$k$ to $k'$: each token is processed by its $k'$ highest-scoring experts, and the remaining $8-k'$ routing slots are dropped. \dpma is held to the same limit, with each token processed by its $k'$ highest-scoring retained experts. Rather than dropping the remaining $8-k'$ slots, \dpma fills them with the highest-scoring absorbed experts whose routing weight is redirected to one of the $k'$ evaluated experts, so these slots adjust the routing weights of the evaluated experts without activating additional ones. We apply this restriction only at evaluation; \dpma is optimised with its original routing. Both \dpma variants outperform every baseline at every tested $k'$, with accuracy averaged over the eight benchmarks and five compression ratios (Figure~\ref{fig:fixed-flops}).

\subsection{Ablations and sensitivity analysis}\label{sec:evaluation-ablations}

We next test whether combining pruning, merging, and absorption improves accuracy, and how \dpma performs when fewer experts are available to compress.
We investigate both questions on OLMoE.

\begin{wraptable}{r}{0.5\textwidth}
\centering
\fontsize{8}{10}\selectfont
\setlength{\tabcolsep}{7pt}
\caption{\textbf{Full \dpma achieves the highest mean accuracy in all four settings.} Accuracy (\%) on OLMoE, averaged over eight multiple-choice benchmarks and five compression ratios.}
\label{tab:operations-summary}
\begin{tabular}{@{}lcccc@{}}
\toprule
 & \multicolumn{2}{c}{\textbf{Zero-shot}} & \multicolumn{2}{c}{\textbf{One-shot}} \\
\cmidrule(lr){2-3}\cmidrule(lr){4-5}
\textbf{Available operations} & \texttt{hard} & \texttt{soft} & \texttt{hard} & \texttt{soft} \\
\midrule
Prune only & 45.7 & 45.7 & 46.0 & 46.0 \\
Prune + merge & 45.5 & 46.5 & 45.8 & 47.0 \\
Prune + absorb & 46.3 & 47.7 & 47.0 & 48.4 \\
\midrule
Full \dpma & \textbf{47.4} & \textbf{48.0} & \textbf{48.1} & \textbf{48.8} \\
\bottomrule
\end{tabular}
\end{wraptable}

\paragraph{Restricting the available operations.}
We optimize \dpma with different subsets of operations, keeping the remaining setup unchanged. Table~\ref{tab:operations-summary} shows that allowing all three operations gives the highest mean accuracy for both variants in zero-shot and one-shot evaluation. In the zero-shot setting, accuracy averaged over the eight benchmarks and five compression ratios rises from $45.7\%$ with pruning alone to $47.4\%$ for full \dpmahard and $48.0\%$ for full \dpmasoft. Adding absorption to pruning improves accuracy in all four settings and provides a larger gain than adding merging alone. Adding merging to pruning and absorption improves accuracy further, although adding it to pruning alone does not improve the hard variant.

\paragraph{Reducing the initial number of experts.}
To test whether \dpma remains effective on models with fewer experts per layer, where it has less room to redistribute the budget across layers and fewer experts to merge or absorb removed ones into, we randomly remove experts from the pretrained OLMoE model without retraining, reducing the number per layer from $64$ to $20$ using three nested draws. We then compress each resulting model at $c=0.50$ and measure its WikiText-2~\citep{wikitext} perplexity after compression divided by its perplexity before compression. Figure~\ref{fig:ppl-ratio-expert-count} reports the geometric mean of this ratio over the three draws. Both \dpma variants achieve lower perplexity ratios than every baseline over most of the tested range, with the gap narrowing at the smallest expert counts. At $20$ experts per layer, $50\%$ compression leaves exactly $10$ experts in every layer, the minimum allowed by \dpma, so no budget can be reallocated across layers.

\section{Conclusion}\label{sec:conclusion}

We introduced \dpma, a retraining-free method that jointly learns compression operations and layer-wise expert allocation under a global expert budget.
By separately controlling parameter merging and routing contributions, it optimizes pruning, merging, and absorption together while keeping the original model weights frozen.
Across four MoE models, \dpma achieves the highest mean accuracy among the evaluated methods, with particularly strong gains under aggressive compression.
We find that learned operation mixtures vary significantly across architectures, while ablations on OLMoE show that pruning, merging, and absorption contribute complementary benefits.
These results answer our central question: \emph{gradient descent can learn both how and where to compress MoE models.}

\section*{Acknowledgments}
We gratefully acknowledge core support from the Science of Fundamental AI Research (SOFAIR) Lab, funded by the Engineering and Physical Sciences Research Council (EPSRC), part of UK Research and Innovation (UKRI), and from the Royal Academy of Engineering through its Chair in Emerging Technologies supporting DANTE. This research also received support from the European Research Council through the REDIAL project; SPRIND through the Composite Learning Challenge; the Foresight Institute through an AI Safety Grant; Google through a Google Academic Research Award; IMEC; and the Ministry of Education of Romania through the Credit and Scholarship Agency; and the Basque Government through the SARETU Elkartek project for basic research (Grant No. KK-2026/00094), the HiTZ-AI excellence research group (IT2004-26), and the IKER-GAITU project; the Ministry of Science, Innovation, and Universities of the Spanish Government (MCIN/AEI/10.13039/501100011033) through the MOLVI project (PID2024-157855OB-C32), co-funded by FEDER, EU, the HumanAIze project (AIA2025-163322-C61), and the CHIST-ERA grant for Project Geo-R2LLM (CHIST-ERA-23-MultiGIS-04; PCI2025-163286); and the European Union under Horizon Europe through Project LUMINOUS (Grant No. 101135724).
\clearpage
\bibliographystyle{\paperbibstyle}
\bibliography{references}

\begin{thebibliography}{31}
\providecommand{\natexlab}[1]{#1}
\providecommand{\url}[1]{\texttt{#1}}
\expandafter\ifx\csname urlstyle\endcsname\relax
  \providecommand{\doi}[1]{doi: #1}\else
  \providecommand{\doi}{doi: \begingroup \urlstyle{rm}\Url}\fi

\bibitem[Bai et~al.(2025)Bai, Li, ZHANG, Hong, and Guo]{diep}
Sikai Bai, Haoxi Li, Jie ZHANG, Zicong Hong, and Song Guo.
\newblock Di{EP}: Adaptive mixture-of-experts compression through differentiable expert pruning.
\newblock In \emph{The Thirty-ninth Annual Conference on Neural Information Processing Systems}, 2025.
\newblock URL \url{https://openreview.net/forum?id=oTocSpfYvJ}.

\bibitem[Bengio et~al.(2013)Bengio, L{\'e}onard, and Courville]{bengio2013estimating}
Yoshua Bengio, Nicholas L{\'e}onard, and Aaron Courville.
\newblock Estimating or propagating gradients through stochastic neurons for conditional computation.
\newblock \emph{arXiv preprint arXiv:1308.3432}, 2013.

\bibitem[Bisk et~al.(2020)Bisk, Zellers, Gao, Choi, et~al.]{piqa}
Yonatan Bisk, Rowan Zellers, Jianfeng Gao, Yejin Choi, et~al.
\newblock Piqa: Reasoning about physical commonsense in natural language.
\newblock In \emph{Proceedings of the AAAI conference on artificial intelligence}, volume~34, pages 7432--7439, 2020.

\bibitem[Chen et~al.(2025)Chen, Liu, Sun, Chao, Hsu, and Lee]{hc_moe}
I-Chun Chen, Hsu-Shen Liu, Wei-Fang Sun, Chen-Hao Chao, Yen-Chang Hsu, and Chun-Yi Lee.
\newblock Retraining-free merging of sparse moe via hierarchical clustering.
\newblock In \emph{Forty-second International Conference on Machine Learning}, 2025.
\newblock URL \url{https://openreview.net/forum?id=hslOzRxzXL}.

\bibitem[Clark et~al.(2019)Clark, Lee, Chang, Kwiatkowski, Collins, and Toutanova]{boolq}
Christopher Clark, Kenton Lee, Ming-Wei Chang, Tom Kwiatkowski, Michael Collins, and Kristina Toutanova.
\newblock {B}ool{Q}: Exploring the surprising difficulty of natural yes/no questions.
\newblock In Jill Burstein, Christy Doran, and Thamar Solorio, editors, \emph{Proceedings of the 2019 Conference of the {N}orth {A}merican Chapter of the {A}ssociation for {C}omputational {L}inguistics: Human Language Technologies, Volume 1 (Long and Short Papers)}, pages 2924--2936, Minneapolis, Minnesota, June 2019. Association for Computational Linguistics.
\newblock \doi{10.18653/v1/N19-1300}.
\newblock URL \url{https://aclanthology.org/N19-1300/}.

\bibitem[Clark et~al.(2018)Clark, Cowhey, Etzioni, Khot, Sabharwal, Schoenick, and Tafjord]{arc}
Peter Clark, Isaac Cowhey, Oren Etzioni, Tushar Khot, Ashish Sabharwal, Carissa Schoenick, and Oyvind Tafjord.
\newblock Think you have solved question answering? try arc, the ai2 reasoning challenge.
\newblock \emph{arXiv preprint arXiv:1803.05457}, 2018.

\bibitem[Dai et~al.(2024)Dai, Deng, Zhao, Xu, Gao, Chen, Li, Zeng, Yu, Wu, et~al.]{dai2024deepseekmoe}
Damai Dai, Chengqi Deng, Chenggang Zhao, RX~Xu, Huazuo Gao, Deli Chen, Jiashi Li, Wangding Zeng, Xingkai Yu, Yu~Wu, et~al.
\newblock Deepseekmoe: Towards ultimate expert specialization in mixture-of-experts language models.
\newblock In \emph{Proceedings of the 62nd annual meeting of the association for computational linguistics (volume 1: Long papers)}, pages 1280--1297, 2024.

\bibitem[Fedus et~al.(2022)Fedus, Zoph, and Shazeer]{fedus2022switch}
William Fedus, Barret Zoph, and Noam Shazeer.
\newblock Switch transformers: Scaling to trillion parameter models with simple and efficient sparsity.
\newblock \emph{Journal of Machine Learning Research}, 23\penalty0 (120):\penalty0 1--39, 2022.

\bibitem[Hendrycks et~al.(2021)Hendrycks, Burns, Basart, Zou, Mazeika, Song, and Steinhardt]{mmlu}
Dan Hendrycks, Collin Burns, Steven Basart, Andy Zou, Mantas Mazeika, Dawn Song, and Jacob Steinhardt.
\newblock Measuring massive multitask language understanding.
\newblock In \emph{International Conference on Learning Representations}, 2021.
\newblock URL \url{https://openreview.net/forum?id=d7KBjmI3GmQ}.

\bibitem[Huang et~al.(2026)Huang, Zhang, Fang, Xiao, and Du]{huang2026prime}
Shanfeng Huang, Rongqian Zhang, Zhou Fang, Song Xiao, and Hai Du.
\newblock Prime: Frequency-aware unified pruning and merging for moe compression.
\newblock \emph{arXiv preprint arXiv:2610.13982}, 2026.

\bibitem[Jacobs et~al.(1991)Jacobs, Jordan, Nowlan, and Hinton]{jacobs1991adaptive}
Robert~A Jacobs, Michael~I Jordan, Steven~J Nowlan, and Geoffrey~E Hinton.
\newblock Adaptive mixtures of local experts.
\newblock \emph{Neural computation}, 3\penalty0 (1):\penalty0 79--87, 1991.

\bibitem[Jaiswal et~al.(2025)Jaiswal, Wang, Li, Li, Chen, Wang, Wang, Pang, and Du]{jaiswal2025finding}
Ajay Jaiswal, Jianyu Wang, Yixiao Li, Pingzhi Li, Tianlong Chen, Zhangyang Wang, Chong Wang, Ruoming Pang, and Xianzhi Du.
\newblock Finding fantastic experts in moes: A unified study for expert dropping strategies and observations.
\newblock \emph{arXiv preprint arXiv:2504.05586}, 2025.

\bibitem[Jha et~al.(2026)Jha, Hashemzadeh, Pasand, Parviz, Lee, and Knyazev]{jha2026ream}
Saurav Jha, Maryam Hashemzadeh, Ali~Saheb Pasand, Ali Parviz, Min-Joong Lee, and Boris Knyazev.
\newblock Ream: Merging improves pruning of experts in llms.
\newblock \emph{arXiv preprint arXiv:2604.04356}, 2026.

\bibitem[Jiang et~al.(2024)Jiang, Sablayrolles, Roux, Mensch, Savary, Bamford, Chaplot, Casas, Hanna, Bressand, et~al.]{jiang2024mixtral}
Albert~Q Jiang, Alexandre Sablayrolles, Antoine Roux, Arthur Mensch, Blanche Savary, Chris Bamford, Devendra~Singh Chaplot, Diego de~las Casas, Emma~Bou Hanna, Florian Bressand, et~al.
\newblock Mixtral of experts.
\newblock \emph{arXiv preprint arXiv:2401.04088}, 2024.

\bibitem[Lasby et~al.(2026)Lasby, Lazarevich, Sinnadurai, Lie, Ioannou, and Thangarasa]{reap}
Mike Lasby, Ivan Lazarevich, Nish Sinnadurai, Sean Lie, Yani Ioannou, and Vithursan Thangarasa.
\newblock {REAP} the experts: Why pruning prevails for one-shot moe compression.
\newblock In \emph{The Fourteenth International Conference on Learning Representations}, 2026.
\newblock URL \url{https://openreview.net/forum?id=ukGxWd2aDG}.

\bibitem[Lepikhin et~al.(2021)Lepikhin, Lee, Xu, Chen, Firat, Huang, Krikun, Shazeer, and Chen]{lepikhin2021gshard}
Dmitry Lepikhin, HyoukJoong Lee, Yuanzhong Xu, Dehao Chen, Orhan Firat, Yanping Huang, Maxim Krikun, Noam Shazeer, and Zhifeng Chen.
\newblock {\{}GS{\}}hard: Scaling giant models with conditional computation and automatic sharding.
\newblock In \emph{International Conference on Learning Representations}, 2021.
\newblock URL \url{https://openreview.net/forum?id=qrwe7XHTmYb}.

\bibitem[Li et~al.(2026)Li, Zhu, Wang, Qin, Li, Gu, Han, and Guo]{li2026submoe}
Lujun Li, Qiyuan Zhu, Jiacheng Wang, Xiaoyu Qin, Wei Li, Hao Gu, Sirui Han, and Yike Guo.
\newblock Sub-moe: Efficient mixture-of-expert llms compression via subspace expert merging.
\newblock In \emph{Proceedings of the AAAI Conference on Artificial Intelligence}, volume~40, pages 22994--23002, 2026.

\bibitem[Li et~al.(2024)Li, Zhang, Yadav, Sung, Cheng, Bansal, and Chen]{mc_smoe}
Pingzhi Li, Zhenyu Zhang, Prateek Yadav, Yi-Lin Sung, Yu~Cheng, Mohit Bansal, and Tianlong Chen.
\newblock Merge, then compress: Demystify efficient {SM}oe with hints from its routing policy.
\newblock In \emph{The Twelfth International Conference on Learning Representations}, 2024.
\newblock URL \url{https://openreview.net/forum?id=eFWG9Cy3WK}.

\bibitem[Liu et~al.(2024)Liu, Zhu, Lin, Ning, Blaschko, Yan, Dai, Yang, and Wang]{EEP}
Enshu Liu, Junyi Zhu, Zinan Lin, Xuefei Ning, Matthew~B Blaschko, Shengen Yan, Guohao Dai, Huazhong Yang, and Yu~Wang.
\newblock Efficient expert pruning for sparse mixture-of-experts language models: Enhancing performance and reducing inference costs.
\newblock \emph{arXiv preprint arXiv:2407.00945}, 2024.

\bibitem[Louizos et~al.(2018)Louizos, Welling, and Kingma]{louizos2018learning}
Christos Louizos, Max Welling, and Diederik~P. Kingma.
\newblock Learning sparse neural networks through $l_0$ regularization.
\newblock In \emph{International Conference on Learning Representations}, 2018.
\newblock URL \url{https://openreview.net/forum?id=H1Y8hhg0b}.

\bibitem[Lu et~al.(2024)Lu, Liu, Xu, Zhou, Huang, Zhang, Yan, and Li]{NAEE}
Xudong Lu, Qi~Liu, Yuhui Xu, Aojun Zhou, Siyuan Huang, Bo~Zhang, Junchi Yan, and Hongsheng Li.
\newblock Not all experts are equal: Efficient expert pruning and skipping for mixture-of-experts large language models.
\newblock In Lun-Wei Ku, Andre Martins, and Vivek Srikumar, editors, \emph{Proceedings of the 62nd Annual Meeting of the {A}ssociation for {C}omputational {L}inguistics (Volume 1: Long Papers)}, pages 6159--6172, Bangkok, Thailand, August 2024. Association for Computational Linguistics.
\newblock \doi{10.18653/v1/2024.acl-long.334}.
\newblock URL \url{https://aclanthology.org/2024.acl-long.334/}.

\bibitem[Merity et~al.(2017)Merity, Xiong, Bradbury, and Socher]{wikitext}
Stephen Merity, Caiming Xiong, James Bradbury, and Richard Socher.
\newblock Pointer sentinel mixture models.
\newblock In \emph{International Conference on Learning Representations}, 2017.
\newblock URL \url{https://openreview.net/forum?id=Byj72udxe}.

\bibitem[Mihaylov et~al.(2018)Mihaylov, Clark, Khot, and Sabharwal]{bookqa}
Todor Mihaylov, Peter Clark, Tushar Khot, and Ashish Sabharwal.
\newblock Can a suit of armor conduct electricity? a new dataset for open book question answering.
\newblock In Ellen Riloff, David Chiang, Julia Hockenmaier, and Jun{'}ichi Tsujii, editors, \emph{Proceedings of the 2018 Conference on Empirical Methods in Natural Language Processing}, pages 2381--2391, Brussels, Belgium, October-November 2018. Association for Computational Linguistics.
\newblock \doi{10.18653/v1/D18-1260}.
\newblock URL \url{https://aclanthology.org/D18-1260/}.

\bibitem[Raffel et~al.(2020)Raffel, Shazeer, Roberts, Lee, Narang, Matena, Zhou, Li, and Liu]{c4}
Colin Raffel, Noam Shazeer, Adam Roberts, Katherine Lee, Sharan Narang, Michael Matena, Yanqi Zhou, Wei Li, and Peter~J. Liu.
\newblock Exploring the limits of transfer learning with a unified text-to-text transformer.
\newblock \emph{Journal of Machine Learning Research}, 21\penalty0 (140):\penalty0 1--67, 2020.

\bibitem[Rajbhandari et~al.(2022)Rajbhandari, Li, Yao, Zhang, Aminabadi, Awan, Rasley, and He]{rajbhandari2022deepspeed}
Samyam Rajbhandari, Conglong Li, Zhewei Yao, Minjia Zhang, Reza~Yazdani Aminabadi, Ammar~Ahmad Awan, Jeff Rasley, and Yuxiong He.
\newblock Deepspeed-moe: Advancing mixture-of-experts inference and training to power next-generation ai scale.
\newblock In \emph{International conference on machine learning}, pages 18332--18346. PMLR, 2022.

\bibitem[Sakaguchi et~al.(2021)Sakaguchi, Bras, Bhagavatula, and Choi]{winogrande}
Keisuke Sakaguchi, Ronan~Le Bras, Chandra Bhagavatula, and Yejin Choi.
\newblock Winogrande: An adversarial winograd schema challenge at scale.
\newblock \emph{Communications of the ACM}, 64\penalty0 (9):\penalty0 99--106, 2021.

\bibitem[Shazeer et~al.(2017)Shazeer, Mirhoseini, Maziarz, Davis, Le, Hinton, and Dean]{shazeer2017outrageously}
Noam Shazeer, *Azalia Mirhoseini, *Krzysztof Maziarz, Andy Davis, Quoc Le, Geoffrey Hinton, and Jeff Dean.
\newblock Outrageously large neural networks: The sparsely-gated mixture-of-experts layer.
\newblock In \emph{International Conference on Learning Representations}, 2017.
\newblock URL \url{https://openreview.net/forum?id=B1ckMDqlg}.

\bibitem[Yang et~al.(2025)Yang, Li, Yang, Zhang, Hui, Zheng, Yu, Gao, Huang, Lv, et~al.]{yang2025qwen3}
An~Yang, Anfeng Li, Baosong Yang, Beichen Zhang, Binyuan Hui, Bo~Zheng, Bowen Yu, Chang Gao, Chengen Huang, Chenxu Lv, et~al.
\newblock Qwen3 technical report.
\newblock \emph{arXiv preprint arXiv:2505.09388}, 2025.

\bibitem[Yao et~al.(2026)Yao, Pan, Dai, Cong, Li, and Yang]{yao2026conmoe}
Yilun Yao, Jiaming Pan, Elsie Dai, Peizhuang Cong, Yaoming Li, and Tong Yang.
\newblock Conmoe: Expert-pool consolidation via prototype reassignment for moe compression.
\newblock \emph{arXiv preprint arXiv:2605.29350}, 2026.

\bibitem[Zellers et~al.(2019)Zellers, Holtzman, Bisk, Farhadi, and Choi]{hellaswag}
Rowan Zellers, Ari Holtzman, Yonatan Bisk, Ali Farhadi, and Yejin Choi.
\newblock {H}ella{S}wag: Can a machine really finish your sentence?
\newblock In Anna Korhonen, David Traum, and Llu{\'i}s M{\`a}rquez, editors, \emph{Proceedings of the 57th Annual Meeting of the {A}ssociation for {C}omputational {L}inguistics}, pages 4791--4800, Florence, Italy, July 2019. Association for Computational Linguistics.
\newblock \doi{10.18653/v1/P19-1472}.
\newblock URL \url{https://aclanthology.org/P19-1472/}.

\bibitem[Zhang et~al.(2026)Zhang, Yuxuan, Lou, Zhang, Zhao, and You]{zhang2026mone}
Geng Zhang, Han Yuxuan, Yuxuan Lou, Yiqi Zhang, Wangbo Zhao, and Yang You.
\newblock Mone: Replacing redundant experts with lightweight novices for structured pruning of moe.
\newblock In \emph{International Conference on Learning Representations}, volume 2026, pages 39983--40009, 2026.

\end{thebibliography}
\clearpage
\appendix
\section{Operations chosen by \dpma to compress the models}\label{app:comp-operations}

We present a summary of the operations that have been selected by \dpma to compress the models used on the experiments on each one of the compression ratios in Table~\ref{tab:operations-chosen-compression}.

\begin{table*}[h]
\centering
\scriptsize
\caption{Operations \dpma applies to the experts it drops, as a percentage of the dropped experts.
Mean $\pm$ the standard deviation over seeds is shown.}
\label{tab:operations-chosen-compression}
\begin{tabular}{llccccc}
\toprule
 & & \multicolumn{5}{c}{\textbf{Compression ratio}} \\
\cmidrule(lr){3-7}
\textbf{Model} & \textbf{Operation} & \textbf{$25\%$} & \textbf{$40\%$} & \textbf{$50\%$} & \textbf{$60\%$} & \textbf{$75\%$} \\
\midrule
\multirow{5}{*}{\textbf{OLMoE}}
 & Experts dropped (of 1{,}024) & 256 & 410 & 512 & 614 & 768 \\
\cmidrule(lr){2-7}
 & Prune          & 7.2\pmstd{2.4} & 7.7\pmstd{1.6} & 8.4\pmstd{0.5} & 10.9\pmstd{0.8} & 10.7\pmstd{0.9} \\
 & Merge          & 5.9\pmstd{1.2} & 7.3\pmstd{0.2} & 7.4\pmstd{0.5} & 8.8\pmstd{1.2} & 9.5\pmstd{2.2} \\
 & Absorb         & 45.2\pmstd{1.5} & 41.0\pmstd{2.5} & 42.0\pmstd{2.8} & 43.2\pmstd{2.8} & 46.4\pmstd{1.0} \\
 & Merge + absorb & 41.8\pmstd{3.1} & 44.0\pmstd{3.7} & 42.2\pmstd{3.1} & 37.1\pmstd{2.9} & 33.4\pmstd{3.4} \\
\midrule
\multirow{5}{*}{\textbf{Qwen1.5-MoE}}
 & Experts dropped (of 1{,}440) & 360 & 576 & 720 & 864 & 1{,}080 \\
\cmidrule(lr){2-7}
 & Prune          & 10.5\pmstd{2.1} & 11.6\pmstd{2.1} & 12.4\pmstd{2.0} & 16.8\pmstd{1.7} & 21.6\pmstd{2.7} \\
 & Merge          & 12.2\pmstd{0.7} & 17.7\pmstd{6.7} & 16.5\pmstd{5.1} & 21.5\pmstd{2.8} & 17.5\pmstd{2.2} \\
 & Absorb         & 31.9\pmstd{2.5} & 33.6\pmstd{2.8} & 36.6\pmstd{2.0} & 37.8\pmstd{0.8} & 39.0\pmstd{1.6} \\
 & Merge + absorb & 45.4\pmstd{1.3} & 37.1\pmstd{5.6} & 34.5\pmstd{6.0} & 24.0\pmstd{3.7} & 21.9\pmstd{2.9} \\
\midrule
\multirow{5}{*}{\textbf{Qwen3}}
 & Experts dropped (of 6{,}144) & 1{,}536 & 2{,}458 & 3{,}072 & 3{,}686 & 4{,}608 \\
\cmidrule(lr){2-7}
 & Prune          & 95.3\pmstd{2.8} & 88.8\pmstd{0.5} & 85.9\pmstd{1.3} & 86.7\pmstd{2.5} & 84.1\pmstd{3.0} \\
 & Merge          & 4.1\pmstd{2.9} & 8.0\pmstd{0.5} & 8.4\pmstd{0.4} & 7.5\pmstd{0.3} & 6.1\pmstd{0.5} \\
 & Absorb         & 0.7\pmstd{0.1} & 3.0\pmstd{0.1} & 5.4\pmstd{0.8} & 5.5\pmstd{1.9} & 9.3\pmstd{2.2} \\
 & Merge + absorb & 0.0\pmstd{0.0} & 0.2\pmstd{0.1} & 0.4\pmstd{0.1} & 0.4\pmstd{0.3} & 0.5\pmstd{0.2} \\
 \midrule
\multirow{5}{*}{\textbf{DeepSeek-MoE}}
 & Experts dropped (of 1{,}728) & 432 & 691 & 864 & 1{,}037 & 1{,}296 \\
\cmidrule(lr){2-7}
 & Prune          & 15.7\pmstd{1.7} & 17.7\pmstd{2.9} & 21.0\pmstd{3.5} & 26.4\pmstd{13.0} & 32.0\pmstd{10.6} \\
 & Merge          & 12.2\pmstd{6.1} & 14.4\pmstd{4.1} & 19.0\pmstd{6.1} & 21.2\pmstd{12.1} & 22.8\pmstd{5.2} \\
 & Absorb         & 29.9\pmstd{7.1} & 30.4\pmstd{1.2} & 29.1\pmstd{2.7} & 27.0\pmstd{9.6} & 27.6\pmstd{6.5} \\
 & Merge + absorb & 42.2\pmstd{3.4} & 37.4\pmstd{6.2} & 30.9\pmstd{8.4} & 25.4\pmstd{15.4} & 17.6\pmstd{9.0} \\
\bottomrule
\end{tabular}
\end{table*}

\begin{figure}[h]
    \centering
    \includegraphics[width=\linewidth]{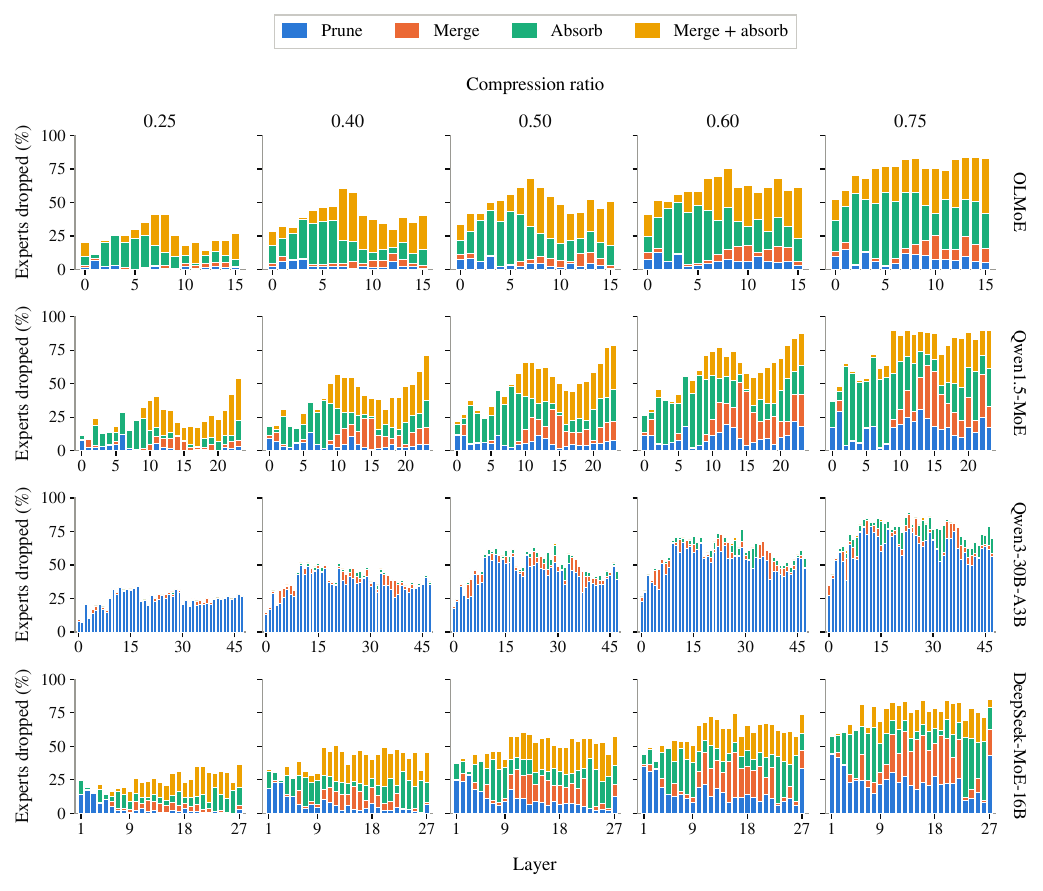}
    \caption{\textbf{Operations \dpma applies per layer.} Experts dropped per MoE layer (\% of the layer's routed experts), stacked by operation, per model (rows) and compression ratio (columns). We represent the mean over three seeds.}
    \label{fig:operations-per-layer}
\end{figure}

As we point out in Section~\ref{sec:compression-recipes}, \dpma's preferred operations are model- and scenario-dependent. Figure~\ref{fig:operations-per-layer} breaks the operations down by layer. Two patterns stand out. First, each model has a recognizable mix that holds across depth and compression ratios: \dpma prunes almost exclusively on Qwen3-30B-A3B, whereas on OLMoE, Qwen1.5-MoE and DeepSeek-MoE-16B it combines all four operations, in line with the model-level shares in Table~\ref{tab:operations-chosen-compression}. Second, within these three models the mix changes with depth. In the first layers, removed experts are mostly absorbed or pruned. Further into the network, merging, alone or together with absorption, becomes more common. \dpma therefore adapts the operation not only to the model but also to the layer.

\section{\dpma's budget allocation}
\begin{figure}
    \centering
    \includegraphics[width=\linewidth]{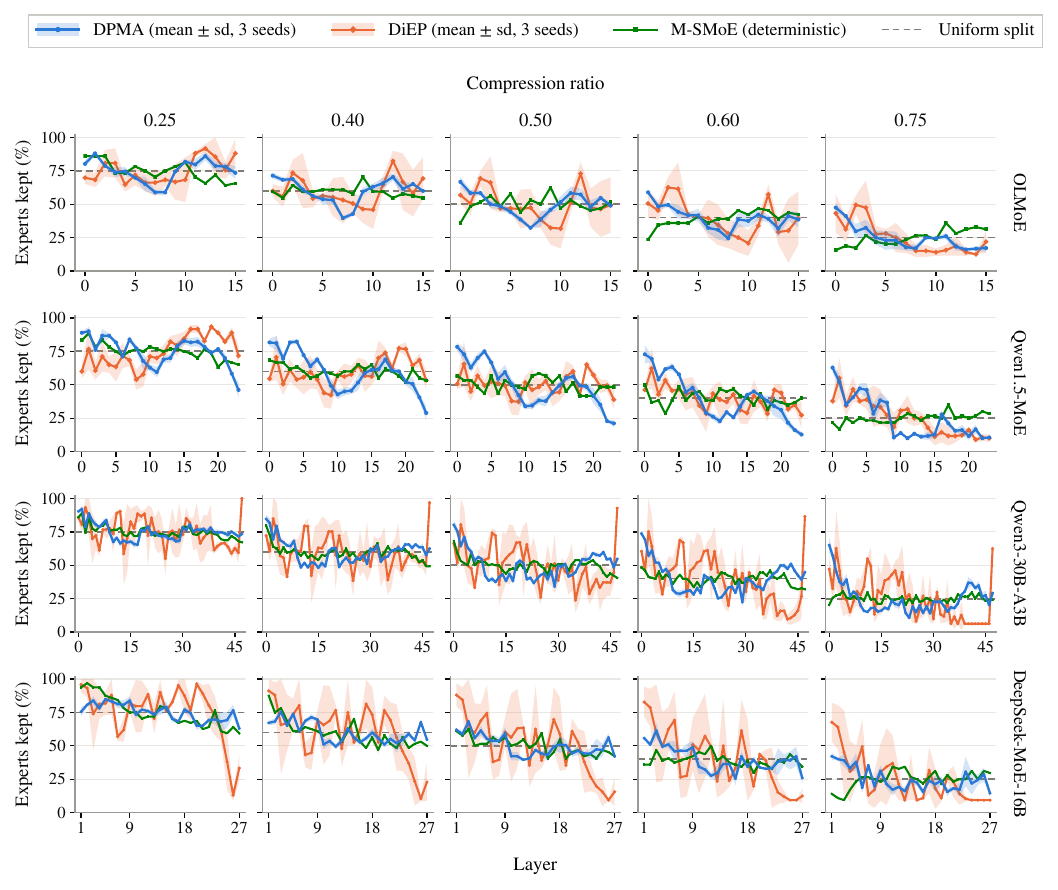}
    \caption{Per-layer budget of \dpma, \diep and \msmoe. Experts kept per MoE layer (\% of the layer's routed experts), per model (rows) and compression ratio (columns). \dpma and \diep: mean over three seeds; \msmoe: single deterministic run.}
    \label{fig:budet-split-all-methods}
\end{figure}

While Section~\ref{sec:results} analyzes overall budget allocation and summarizes the mean compression patterns chosen by \dpma (Figure~\ref{fig:expert_budget_allocation}), Figure~\ref{fig:budet-split-all-methods} provides the fine-grained, per-layer breakdown across all models. Specifically, it compares the layer-wise splits discovered by \dpma against those of \diep and \msmoe, the baselines capable of dynamically adjusting per-layer expert budgets under a global constraint.

\section{\dpmahard vs \dpmasoft}\label{app:hard-vs-soft}

\dpmahard and \dpmasoft derive from the same optimization and differ only in how the assignment matrix $A_{er}$ is applied at export. This choice only affects experts that are absorbed, alone or together with merging. Pruned and merged experts are removed from the router. An absorbed expert, in contrast, keeps its routing slot, and the routing weight it receives is passed on to surviving experts: \dpmahard passes it to the single most likely target in $A_{er}$, while \dpmasoft splits it over the top-3 targets. The main consequence is the number of experts that process each token, which Table~\ref{tab:routing-compute-c4} reports per layer for every model and compression ratio.

Under \dpmahard, a token never uses more experts than the original top-$k$. An absorbed slot is redirected to an expert that may already be among the token's selected experts, in which case both contributions fall on the same expert, so tokens use slightly fewer experts on average, increasingly so as compression grows. Under \dpmasoft, the weight of an absorbed slot can reach experts outside the token's original selection, so a token may use more experts than the original top-$k$.

How far the two variants diverge depends on the operations \dpma selects (Table~\ref{tab:operations-chosen-compression}). On Qwen3-30B-A3B, \dpma prunes almost every removed expert, so few routing slots are absorbed and both variants use essentially the original number of experts per token. On OLMoE, Qwen1.5-MoE and DeepSeek-MoE-16B, absorption is a large part of the mix, and \dpmasoft uses more experts per token than \dpmahard. This overhead does not grow in proportion to the compression ratio. The additional experts that \dpmasoft activates are always drawn from the surviving experts, and the pool of survivors shrinks as compression increases. With mild compression, survivors are plentiful, and the targets of an absorbed slot are usually experts the token is not already using. With strong compression, many absorbed slots point to the same few survivors, which increasingly coincide with each other and with the experts the token already selected, so each absorbed slot adds fewer new experts and the overhead levels off. A single training run thus yields two operating points: \dpmahard keeps the number of experts per token at or below that of the original model, while \dpmasoft uses additional experts where \dpma has chosen to absorb, and is slightly more accurate in every setting of Table~\ref{tab:main-results-0shot}.

\begin{table}[h]
\centering
\scriptsize
\caption{Experts activated per token under soft vs.\ hard routing.
$B$: routed experts kept, for compression ratios $25\%$, $40\%$, $50\%$, $60\%$, and $75\%$ (top to bottom). Params vs.\ original: parameters of the compressed model, w.r.t the uncompressed counterpart. Experts/token: routed experts each token activates, measured on held-out C4 sequences (a total of 4096 tokens). $\pm$ is the standard deviation over 3 seeds.}
\label{tab:routing-compute-c4}
\begin{tabular}{l cc cc}
\toprule
& & & \multicolumn{2}{c}{Experts/token} \\
\cmidrule(lr){4-5}
Model & $B$ & Params vs.\ original & Hard & Soft \\
\midrule
\multirow{5}{*}{\makecell[l]{\textbf{OLMoE}\\$k=8$}} & 768 & 76.7\% & 7.70\,{\scriptsize$\pm$0.02} & 8.84\,{\scriptsize$\pm$0.02} \\
 & 614 & 62.7\% & 7.38\,{\scriptsize$\pm$0.02} & 9.20\,{\scriptsize$\pm$0.08} \\
 & 512 & 53.4\% & 7.07\,{\scriptsize$\pm$0.06} & 9.27\,{\scriptsize$\pm$0.03} \\
 & 410 & 44.2\% & 6.73\,{\scriptsize$\pm$0.02} & 9.17\,{\scriptsize$\pm$0.06} \\
 & 256 & 30.2\% & 5.78\,{\scriptsize$\pm$0.06} & 8.26\,{\scriptsize$\pm$0.09} \\
\midrule
\multirow{5}{*}{\makecell[l]{\textbf{Qwen1.5-MoE}\\$k=4$}} & 1{,}080 & 78.2\% & 3.89\,{\scriptsize$\pm$0.00} & 4.54\,{\scriptsize$\pm$0.06} \\
 & 864 & 65.2\% & 3.82\,{\scriptsize$\pm$0.05} & 4.78\,{\scriptsize$\pm$0.24} \\
 & 720 & 56.5\% & 3.72\,{\scriptsize$\pm$0.05} & 5.04\,{\scriptsize$\pm$0.14} \\
 & 576 & 47.8\% & 3.67\,{\scriptsize$\pm$0.04} & 4.99\,{\scriptsize$\pm$0.14} \\
 & 360 & 34.7\% & 3.41\,{\scriptsize$\pm$0.03} & 5.04\,{\scriptsize$\pm$0.02} \\
\midrule
\multirow{5}{*}{\makecell[l]{\textbf{Qwen3}\\$k=8$}} & 4{,}608 & 76.3\% & 8.00\,{\scriptsize$\pm$0.00} & 8.00\,{\scriptsize$\pm$0.00} \\
 & 3{,}686 & 62.0\% & 8.00\,{\scriptsize$\pm$0.00} & 8.00\,{\scriptsize$\pm$0.00} \\
 & 3{,}072 & 52.5\% & 7.99\,{\scriptsize$\pm$0.00} & 8.02\,{\scriptsize$\pm$0.01} \\
 & 2{,}458 & 43.0\% & 7.99\,{\scriptsize$\pm$0.01} & 8.03\,{\scriptsize$\pm$0.02} \\
 & 1{,}536 & 28.8\% & 7.90\,{\scriptsize$\pm$0.04} & 8.16\,{\scriptsize$\pm$0.06} \\
\midrule
\multirow{5}{*}{\makecell[l]{\textbf{DeepSeek-MoE}\\$k=6$}} & 1{,}296 & 77.2\% & 5.85\,{\scriptsize$\pm$0.03} & 6.65\,{\scriptsize$\pm$0.11} \\
 & 1{,}037 & 63.5\% & 5.73\,{\scriptsize$\pm$0.01} & 6.96\,{\scriptsize$\pm$0.11} \\
 & 864 & 54.4\% & 5.66\,{\scriptsize$\pm$0.05} & 7.06\,{\scriptsize$\pm$0.22} \\
 & 691 & 45.2\% & 5.57\,{\scriptsize$\pm$0.25} & 7.01\,{\scriptsize$\pm$0.63} \\
 & 432 & 31.5\% & 5.29\,{\scriptsize$\pm$0.29} & 6.80\,{\scriptsize$\pm$0.33} \\
\bottomrule
\end{tabular}
\end{table}

\section{Extra ablation studies and sensitivity analysis}\label{app:ablations-sensitivity}
We further analyze how key components affect \dpma, examining the impact of calibration token count, the Lagrangian term enforcing budget completion, and the layer-wise loss.

\subsection{Importance of the number of calibration tokens}

\begin{figure}[h]
    \centering
    \includegraphics[width=\linewidth]{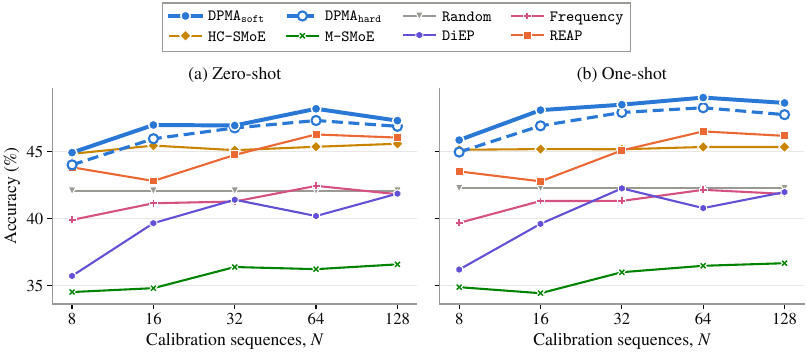}
    \caption{Effect of the calibration set size on OLMoE at compression ratio of $50\%$. Mean accuracy (\%) over 8 MC tasks, using $N$ C4 sequences of 1{,}024 tokens; the main experiments use $N{=}32$. $\pm$ is the standard deviation over seeds (\diep, \dpma) or random draws (\rand).}
    \label{fig:calibration-size}
\end{figure}

Figure~\ref{fig:calibration-size} varies the number of calibration sequences on OLMoE at a compression of $50\%$. \dpma is the most accurate method at every calibration size in both settings: \dpmasoft at every size, \dpmahard from 16 sequences on. With only 16 sequences, \dpmasoft already outperforms every baseline, including baselines given 128 sequences, and from 32 sequences on, the size used in our main experiments, its accuracy levels off. \diep, the other method that learns from the calibration data, loses at least twice as much accuracy as \dpma when the calibration set shrinks to 8 sequences. Learning a compression plan therefore does not require a large calibration set.

\subsection{The lagrangian term}
During the soft phase, the augmented Lagrangian is what pushes the number of open experts towards the budget $B$. To measure its effect, we remove it, setting both the multiplier $\mu$ and the penalty weight $\kappa$ to zero, and keep everything else unchanged. The budget is then enforced only at the switch to the hard phase, where the $B$ experts with the largest keep gates are retained. The hard phase then proceeds as usual, so the repairs are still trained on the retained experts.

The term does what it is designed for. With it, the number of open experts approaches the budget during the soft phase, and the projection at the switch only needs to re-admit a small number of experts. Without it, nothing encourages the gates to close: virtually every expert remains open until the switch, and the projection removes all of them at once, based on gate values that were never pressured to discriminate between experts.

Table~\ref{tab:lagrangian} shows the effect on accuracy. Removing the term lowers the accuracy of both \dpma variants at every compression ratio from $40\%$ upwards, in both the zero-shot and the one-shot setting. The exception is the $25\%$ compression ratio, where removing the term improves both variants and yields the best zero-shot accuracy of any method at this ratio. Without the term, \dpma essentially keeps the saliency ranking used to initialize the gates, and we hypothesize that when only a few experts are removed, this criterion can suffice: pruning by saliency alone, without any training, already outperforms the full \dpma at this ratio. As more experts are removed, it clearly does not, falling well behind \dpma from a compression ratio of $50\%$ upwards, and the selection learned under the budget pressure of the soft phase translates into higher accuracy.

\begin{table*}[h]
\centering
\scriptsize
\caption{Ablation of the Lagrangian budget term on OLMoE.
Mean accuracy over 8 MC tasks; $\pm$ is the standard deviation over 3 seeds.
\emph{w/o Lagrangian} removes the augmented Lagrangian from the soft phase ($\mu \equiv 0$, $\kappa = 0$), so the budget is met only by the projection to the top-$B$ experts at the switch to hard gates; everything else, including the initialization, is unchanged.
Bold: the better of each pair.}
\label{tab:lagrangian}
\begin{tabular}{llccccc}
\toprule
 & & \multicolumn{5}{c}{\textbf{Compression ratio}} \\
\cmidrule(lr){3-7}
\textbf{Setting} & \textbf{Method} & \textbf{25\%} & \textbf{40\%} & \textbf{50\%} & \textbf{60\%} & \textbf{75\%} \\
\midrule
\multirow{4}{*}{\textbf{Zero-shot}}
 & \dpmahard & 57.1\pmstd{0.7} & \textbf{51.5}\pmstd{1.2} & \textbf{46.8}\pmstd{0.4} & \textbf{43.0}\pmstd{2.2} & \textbf{38.9}\pmstd{0.1} \\
 & \quad w/o Lagrangian & \textbf{59.0}\pmstd{0.5} & 50.3\pmstd{0.7} & 45.3\pmstd{0.5} & 41.2\pmstd{1.0} & 37.2\pmstd{0.6} \\
\cmidrule(lr){2-7}
 & \dpmasoft & 57.9\pmstd{0.5} & \textbf{52.1}\pmstd{1.3} & \textbf{46.9}\pmstd{1.0} & \textbf{43.5}\pmstd{1.8} & \textbf{39.5}\pmstd{0.8} \\
 & \quad w/o Lagrangian & \textbf{59.1}\pmstd{0.6} & 50.2\pmstd{0.7} & 45.9\pmstd{0.1} & 41.6\pmstd{1.2} & 38.0\pmstd{1.0} \\
\midrule
\multirow{4}{*}{\textbf{One-shot}}
 & \dpmahard & 57.7\pmstd{1.6} & \textbf{52.5}\pmstd{0.1} & \textbf{47.9}\pmstd{0.8} & \textbf{43.6}\pmstd{1.7} & \textbf{38.6}\pmstd{0.1} \\
 & \quad w/o Lagrangian & \textbf{60.0}\pmstd{0.4} & 50.6\pmstd{0.5} & 46.1\pmstd{0.1} & 41.5\pmstd{0.6} & 37.5\pmstd{0.8} \\
\cmidrule(lr){2-7}
 & \dpmasoft & 58.5\pmstd{1.0} & \textbf{53.1}\pmstd{0.8} & \textbf{48.5}\pmstd{0.4} & \textbf{44.7}\pmstd{2.1} & \textbf{39.5}\pmstd{0.3} \\
 & \quad w/o Lagrangian & \textbf{60.0}\pmstd{0.3} & 50.5\pmstd{0.5} & 46.7\pmstd{0.1} & 42.3\pmstd{0.9} & 37.5\pmstd{1.2} \\
\bottomrule
\end{tabular}
\end{table*}

\subsection{The layer-wise loss}
Preserving the outputs of each layer is a well known objective in MoE expert pruning: \naee~\citep{NAEE} selects, layer
by layer, the subset of experts that minimizes the reconstruction error of the layer's output on calibration data.
\dpma uses a related term during the soft phase, the relative error of every layer's hidden states with respect to
the original model, weighted by $\eta$ and optimized jointly with the KL divergence rather than as a per-layer
selection criterion. To assess its role, we retrain \dpma on OLMoE at compression ratio $0.5$ with $\eta$ ranging from $0$
(term removed) to $1.6$, three seeds each.

\begin{figure}[h]
    \centering
    \includegraphics[width=\linewidth]{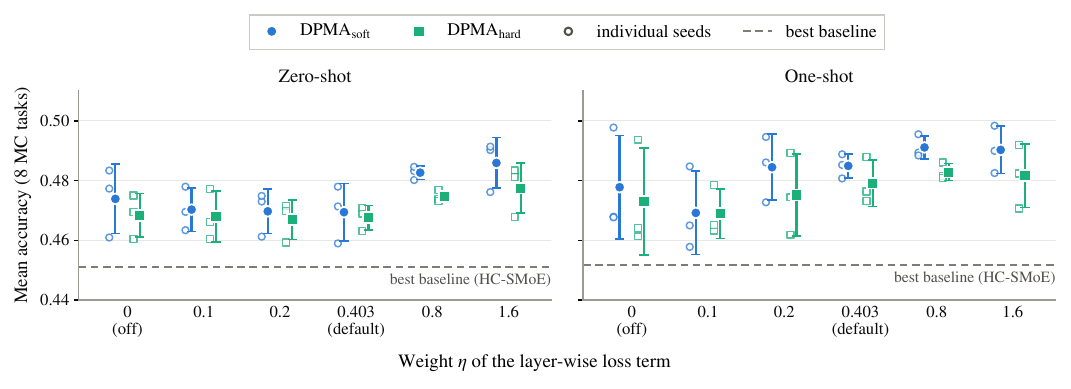}
    \caption{\textbf{Sensitivity of \dpma to the weight $\eta$ of the layer-wise loss.} Mean accuracy over 8 MC tasks on OLMoE at compression ratio $0.50$, zero-shot (left) and one-shot (right). No value of $\eta$ differs significantly from the default.}

    \label{fig:dpma-eta-sensitivity}
\end{figure}

Downstream accuracy does not single out any value of $\eta$ (Figure~\ref{fig:dpma-eta-sensitivity}). With three seeds, the differences between settings stay within the spread across seeds in both the zero-shot and the one-shot setting, and no value differs significantly from the default. 

\begin{figure}[h]
    \centering
    \includegraphics[width=\linewidth]{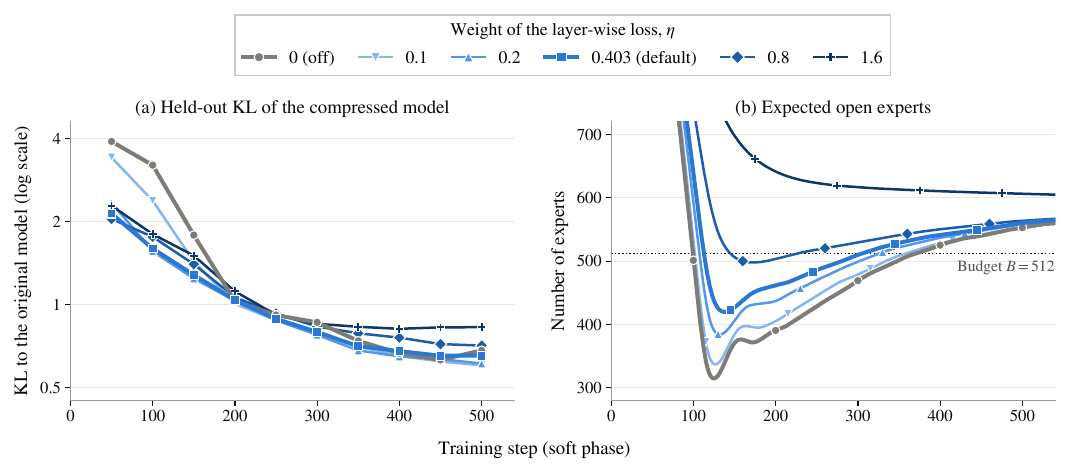}
    \caption{\textbf{The layer-wise loss makes the soft phase close gates gradually.} Soft phase of \dpma on OLMoE at compression ratio 0.5, mean over three seeds per $\eta$. (a) KL divergence to the original model of the compressed model as it would be exported at each step, on held-out C4 windows. (b) Expected number of open experts against the budget $B$. Without the term ($\eta = 0$), gates close faster and early held-out KL is higher; all runs converge to similar values later in training.}
    \label{fig:dpma-eta-dynamics}
\end{figure}

Nevertheless, downstream benchmarks are not the signal $\eta$ should be tuned on, however, since that would tune the compression to the evaluation tasks. The quantities available during training are the held-out KL of the model as it would be exported, and how the gates approach the budget (Figure~\ref{fig:dpma-eta-dynamics}). Both favour keeping the term. Without it ($\eta = 0$), gates close abruptly early in the soft phase: the expected number of open experts drops far below the budget, and the held-out KL of the exported model is about twice as high early on. As $\eta$ grows, gates close more gradually. At the largest values, however, they no longer come down to the budget during the soft phase, leaving the projection at the switch to close the remaining gap, and the held-out KL of the selected checkpoint gets worse. Intermediate values, including the default, avoid both effects. We therefore keep the term with a moderate weight, and treat it as a regularizer of the soft phase rather than a source of accuracy gains.

\section{Overhead of \dpma.} \dpma adds very few parameters and never changes the model's weights. What it learns
is the gate logits, the merge magnitudes and the metric factor: between 20{,}480 and 40{,}960 numbers depending on
the model (see Table~\ref{tab:dpma-overhead}). Because the base model stays frozen, the original model used as the teacher and the
model being compressed share the same weights. Only one copy of the model is held in memory, so every run fits on a single GPU, including Qwen3-30B-A3B. None of these parameters is needed after export.

Learning a compression plan takes about 4 to 8 GPU hours per model and compression ratio (Table~\ref{tab:dpma-overhead}). This is longer than the baselines take. They collect routing or activation
statistics in a single pass over the calibration data, or in the case of \diep run a short differentiable search, and finish within minutes on most models and in about an hour on Qwen3-30B-A3B. The cost of \dpma, however, is paid once, offline, before the model is deployed, and one run yields both \dpma variants. In exchange, \dpma searches jointly over how the budget is split across layers and which operation each removed expert receives, without any fine-tuning of the model's weights, and this is what produces the gains reported in Section~\ref{sec:results}.

\begin{table}[h]
\centering
\scriptsize
\caption{Overhead of \dpma: parameters learned and mean time per run (one model and compression ratio, over 15 runs).}
\label{tab:dpma-overhead}
\begin{tabular}{lcccc}
\toprule
\textbf{Model} & \textbf{Model parameters} & \textbf{Learned parameters by \dpma} & \textbf{Time per run} & \textbf{GPU} \\
\midrule
OLMoE            & 6.9B  & 20{,}480 & 3.8 h & L40S \\
Qwen1.5-MoE      & 14.3B & 22{,}144 & 6.7 h & L40S \\
Qwen3-30B-A3B    & 30.5B & 40{,}960 & 7.0 h & H200 \\
DeepSeek-MoE-16B & 16.4B & 23{,}296 & 8.1 h & L40S \\
\bottomrule
\end{tabular}
\end{table}

\section{Hyperparameters}
\label{app:hyperparameters}

\dpma uses the same configuration for every model and compression ratio (Table~\ref{tab:dpma-hyperparameters}).
The base model stays frozen. Only the gate logits ($\phi_\rho$, $\phi_\alpha$, $\phi_\beta$), the merge magnitudes ($b$) and the low-rank factor $V$ of the metric that defines the direction matrix are learned. The keep-gate logits $\phi_\rho$ are initialized from the \reap saliency of each expert on the calibration data, standardized within each layer, so that every expert starts fully open and more salient experts start further from being closed; the absorb and merge gates start closed. Training has two phases. In the soft phase, the budget is enforced through an augmented Lagrangian. In the hard phase, the kept set is fixed to exactly $B$ experts, with at least top-$k{+}2$ experts in every layer. The optimizer state is reset at the switch between phases. The weight $\eta$ of the layer-wise loss applies only to the soft phase. We set it heuristically on OLMoE, so that at step 300 the weighted layer-wise term is roughly as large as the KL divergence, which keeps either term from dominating early training. This gives one value per compression ratio, and we reuse these values unchanged on every other model, with no further tuning. We keep the checkpoint with the lowest KL divergence on a small held-out C4 set, disjoint from the calibration data, and export both \dpma
variants from it. Every configuration is run with three seeds on a single GPU.

\begin{table}[h]
\centering
\scriptsize
\caption{\dpma hyperparameters, shared by all models.}
\label{tab:dpma-hyperparameters}
\begin{tabular}{ll}
\toprule
\textbf{Setting} & \textbf{Value} \\
\midrule
Calibration data & C4, sequences of 1{,}024 tokens (32; 128 for Qwen3-30B-A3B) \\
Initialization & $\phi_\rho$: \reap saliency, standardized per layer (all experts open) \\
  & $\phi_\alpha$, $\phi_\beta$: closed (no absorption or merging at the start) \\
Optimizer & AdamW ($\beta_1{=}0.9$, $\beta_2{=}0.95$), no weight decay \\
Learning rate & 0.05 for the gate logits ($\phi_\rho$, $\phi_\alpha$, $\phi_\beta$) \\
 & 0.01 for the merge magnitudes ($b$) and the metric factor ($V$) \\
Learning-rate schedule & linear warm-up in each phase; cosine decay in the hard phase \\
Batch size & 4 sequences \\
Training steps & 900: 540 soft phase, 360 hard phase \\
Direction-matrix temperature $\tau$ & cosine anneal: $1.0 \rightarrow 0.3$ over the soft phase, \\
 & \quad $0.3 \rightarrow 0.05$ over the first half of the hard phase, then fixed \\
Layer-wise loss weight $\eta$ & 0.535 / 0.447 / 0.403 / 0.321 / 0.246 \\
 & \quad at compression ratio 0.25 / 0.40 / 0.50 / 0.60 / 0.75 \\
Budget penalty weight $\kappa$ & 100 (soft phase) \\
Lagrange multiplier $\mu$ & adjusted by dual ascent (step size 0.2, starting at 0; soft phase) \\
Gradient clipping & 1.0 \\
Seeds & 0, 1, 2 \\
\bottomrule
\end{tabular}
\end{table}

\section{Hard-concrete gates and non-zero probabilities}\label{app:pnz}
This appendix specifies the hard-concrete gates and derives Equation~\eqref{eq:pnz}, which is the per-gate term of the expected $L_0$ norm of \citet{louizos2018learning}, in our notation.
Throughout, $\sigma(c)=1/(1+e^{-c})$, $\gamma<0<\zeta$, and $t>0$.

\paragraph{Gate construction.}
For a learned logit $\phi$, with $\xi\sim\mathcal{U}(0,1)$ drawn afresh at every forward pass, $\gamma=-0.1$, $\zeta=1.1$, and $t=2/3$, we compute
\begin{equation}
    s=\sigma\!\Bigl(\tfrac{\log\xi-\log(1-\xi)+\phi}{t}\Bigr),
    \qquad
    \mathrm{gate}(\phi)=\operatorname{clip}\bigl(s(\zeta-\gamma)+\gamma,\,0,\,1\bigr).
    \label{eq:gate}
\end{equation}
Stretching the sigmoid output beyond $[0,1]$ and clipping places finite probability mass at both endpoints.
For deterministic evaluation, we use $s=\sigma(\phi)$ and apply the same stretch and clip.

\paragraph{Derivation of non-zero probability.}
Let $s=\sigma(z)$,
$z=(L+\phi^{\rho}_i)/t$ and $L=\log\frac{\xi}{1-\xi}$, where $\xi\sim\mathcal{U}(0,1)$ is
drawn afresh at every forward pass. Only $L$ is random: $\phi^{\rho}_i$ acts as a fixed signal
to which noise is added, and the gate is non-zero when the result is large enough. We first
reduce the event $\rho_i\neq0$ to a threshold condition on $L$, and then use the distribution
of $L$.

The clip maps non-positive values to zero and leaves positive values non-zero. Dividing by
$\zeta-\gamma>0$, and using that $\sigma$ is increasing with inverse
$\operatorname{logit}(c)=\log\frac{c}{1-c}$,
\begin{align*}
    \rho_i\neq0
    &\iff s\,(\zeta-\gamma)+\gamma>0
    \iff \sigma(z)>\frac{-\gamma}{\zeta-\gamma}
    \iff z>\operatorname{logit}\Bigl(\frac{-\gamma}{\zeta-\gamma}\Bigr)=\log\frac{-\gamma}{\zeta}\\
    &\iff L>t\log\frac{-\gamma}{\zeta}-\phi^{\rho}_i=:a,
\end{align*}
where we used $1-\frac{-\gamma}{\zeta-\gamma}=\frac{\zeta}{\zeta-\gamma}$ in the logit, and
$t>0$ in the last step. For $\gamma=-0.1$ and $\zeta=1.1$, the gate is thus non-zero when
$s>1/12$, equivalently when $z>-\log 11\approx-2.398$. The gate is therefore non-zero exactly
when the noise exceeds a bar $a$ that depends only on fixed quantities, and a larger signal
$\phi^{\rho}_i$ lowers the bar.

It remains to compute $\Pr(L>a)$. The map $\xi\mapsto L$ is increasing, and the sigmoid undoes
it,
\begin{equation*}
    \sigma(L)=\frac{1}{1+e^{-L}}=\frac{1}{1+(1-\xi)/\xi}=\xi .
\end{equation*}
Hence $L\le c$ holds exactly when $\xi\le\sigma(c)$, and since $\xi$ is uniform on $(0,1)$,
$\Pr(L\le c)=\sigma(c)$ for every $c\in\mathbb{R}$; that is, $L$ follows a standard logistic
distribution. This sigmoid is unrelated to the one inside the gate: it is the cumulative
distribution function of $L$. Consequently
\begin{equation*}
    \Pr(L>a)=1-\sigma(a)=\frac{e^{-a}}{1+e^{-a}}=\sigma(-a),
\end{equation*}
and combining with the previous paragraph,
\begin{equation}
    P_i=\Pr(\rho_i\neq0)=\sigma(-a)
    =\sigma\Bigl(\phi^{\rho}_i-t\log\frac{-\gamma}{\zeta}\Bigr).
    \label{eq:pnz}
\end{equation}
For $\gamma=-0.1$, $\zeta=1.1$ and $t=2/3$, $-t\log(-\gamma/\zeta)=\tfrac23\log 11\approx1.5986$,
which gives $P_i=\sigma(\phi^{\rho}_i+1.5986)$. The constant is determined by the stretch
interval and the temperature; the noise contributes only the sigmoid shape.

\end{document}